\documentclass[sigconf,nonacm]{acmart} % For LaTeX2e

\usepackage{amsmath}
\usepackage{amsfonts}
\usepackage{bm}
\usepackage{amsthm}
\usepackage{mathtools}
\usepackage[utf8]{inputenc} % allow utf-8 input
\usepackage{hyperref}       % hyperlinks
\usepackage{breakurl}
\usepackage{url}            % simple URL typesetting
\usepackage{booktabs}       % professional-quality tables
\usepackage{nicefrac}       % compact symbols for 1/2, etc.
\usepackage{microtype}      % microtypography
\usepackage[linesnumbered,ruled,vlined]{algorithm2e}
\usepackage{algorithmic}
\usepackage{caption}
\usepackage{graphicx}
\usepackage{tikz}
\usepackage{multirow}
\usepackage{xr}

\makeatletter
\newcommand*{\addFileDependency}[1]{% argument=file name and extension
  \typeout{(#1)}
  \@addtofilelist{#1}
  \IfFileExists{#1}{}{\typeout{No file #1.}}
}
\makeatother
\newcommand*{\myexternaldocument}[1]{%
    \externaldocument{#1}%
    \addFileDependency{#1.tex}%
    \addFileDependency{#1.aux}%
}

\usetikzlibrary{shapes.geometric}

\newcommand{\obs}{\sO}
\newcommand{\batch}{\sB}
\newcommand{\batchb}{\hat{\obs}_1}
\newcommand{\batchc}{\hat{\obs}_2}
\newcommand{\batchh}{\hat{\obs}}

\newcommand{\regularizer}{\ensuremath{\Psi}}
\newcommand{\Wu}{\ensuremath{\mA}}
\newcommand{\Wv}{\ensuremath{\mB}}
\newcommand{\wi}{\ensuremath{a_i}}
\newcommand{\wj}{\ensuremath{b_j}}
\newcommand{\nw}{\ensuremath{\omega}}

\newcommand{\gwi}{\ensuremath{\bar{a}_i}}
\newcommand{\gwj}{\ensuremath{\bar{b}_j}}

\newcommand{\CG}{\text{\#CG}\xspace}
\newcommand{\LS}{\text{\#LS}\xspace}

\newcommand{\ssum}{\ensuremath{\sum_{(i, j) \in \obs}}}
\newcommand{\bsumb}{\ensuremath{\sum_{(i, j) \in \batchb}}}
\newcommand{\bsumc}{\ensuremath{\sum_{(i, j) \in \batchc}}}

\newcommand{\issum}{\ensuremath{\sum_{(i, j') \in \obs}}}
\newcommand{\jssum}{\ensuremath{\sum_{(i', j) \in \obs}}}
\newcommand{\msum}{\ensuremath{\sum_{i=1}^m}}
\newcommand{\nsum}{\ensuremath{\sum_{j=1}^n}}
\newcommand{\dbsum}{\ensuremath{\msum\nsum}}

\newcommand{\csum}{\ensuremath{\sum\nolimits}}
\newcommand{\cmsum}{\ensuremath{\csum_{i=1}^m}}
\newcommand{\cnsum}{\ensuremath{\csum_{j=1}^n}}
\newcommand{\cdbsum}{\ensuremath{\cmsum\cnsum}}

\newcommand{\Pc}{\mP_c}
\newcommand{\Qc}{\mQ_c}
\newcommand{\tPc}{\tilde{\mP}_c}
\newcommand{\tQc}{\tilde{\mQ}_c}
\newcommand{\hPc}{\hat{\mP}_c}
\newcommand{\hQc}{\hat{\mQ}_c}

\newcommand{\Po}{\mP_{\obs}}
\newcommand{\Qo}{\mQ_{\obs}}
\newcommand{\hPo}{\hat{\mP}_{\obs}}
\newcommand{\hQo}{\hat{\mQ}_{\obs}}

\newcommand{\Pbb}{\mP_{1}}
\newcommand{\Qbb}{\mQ_{1}}
\newcommand{\hPbb}{\hat{\mP}_{1}}
\newcommand{\hQbb}{\hat{\mQ}_{1}}

\newcommand{\Pbc}{\mP_{2}}
\newcommand{\Qbc}{\mQ_{2}}
\newcommand{\hPbc}{\hat{\mP}_{2}}
\newcommand{\hQbc}{\hat{\mQ}_{2}}

\newcommand{\bwi}{\ensuremath{\vw_i}}
\newcommand{\bhj}{\ensuremath{\vh_j}}

\newcommand{\hYij}{\hat{Y}_{ij}}
\newcommand{\tYij}{\tilde{Y}_{ij}}
\newcommand{\Yij}{Y_{ij}}

\newcommand{\Jyij}{\parti{\hYij}{\vtheta}}
\newcommand{\xJyij}{\xparti{\hYij}{\vtheta}}
\newcommand{\Jfi}{\parti{f^i}{\vtheta^u}}
\newcommand{\xJfi}{\xparti{f^i}{\vtheta^u}}
\newcommand{\Jgj}{\parti{g^j}{\vtheta^v}}
\newcommand{\xJgj}{\xparti{g^j}{\vtheta^v}}
\newcommand{\Jyf}{\parti{\hYij}{f^i}}
\newcommand{\Jyg}{\parti{\hYij}{g^j}}

\newcommand{\Tp}[1]{#1^\top}
\newcommand{\tJyij}{\Tp{\Jyij}}
\newcommand{\tJfi}{\Tp{\Jfi}}
\newcommand{\tJgj}{\Tp{\Jgj}}

\newcommand{\tJf}{\ensuremath{\tJ^u}}
\newcommand{\tJg}{\ensuremath{\tJ^v}}

\newcommand{\Zi}{\ensuremath{{\vp}_i}}
\newcommand{\Zj}{\ensuremath{{\vq}_j}}

\newcommand{\tZi}{\ensuremath{\tilde{{\vp}}_i}}
\newcommand{\tZj}{\ensuremath{\tilde{{\vq}}_j}}

\newcommand{\Fi}{\ensuremath{f}^{i}(\vtheta^u)}
\newcommand{\Fj}{\ensuremath{g}^{j}(\vtheta^v)}

\newcommand{\fip}[2]{\ensuremath{\langle#1,#2 \rangle_\text{F}}}

\newcommand{\La}{\ensuremath{L(\vtheta)}}
\newcommand{\Lp}{\ensuremath{L^+(\vtheta)}}
\newcommand{\Ln}{\ensuremath{L^-(\vtheta)}}

\newcommand{\lij}{\ell_{ij}}

\newcommand{\diag}{\ensuremath{\text{diag}}}

\newcommand{\half}{\ensuremath{\frac{1}{2}}}
\newcommand{\bbO}[1]{\ensuremath{\mathcal{O}(#1)}}

\newcommand{\grad} {\ensuremath {\nabla {L}(\vtheta)}}
\newcommand{\pgrad} {\ensuremath {\nabla {L}^{+}(\vtheta)}}
\newcommand{\ngrad} {\ensuremath {\nabla {L}^{-}(\vtheta)}}

\newcommand{\hess} {\ensuremath {\nabla^2 {L}(\vtheta)}}

\newcommand{\parti}[2]{\ensuremath{\frac{\partial #1}{\partial #2}}}
\newcommand{\sparti}[2]{\ensuremath{\frac{\partial^2 #1}{{\partial {#2}^2}}}}
\newcommand{\xparti}[2]{\ensuremath{\partial #1 / \partial #2}}

\newcommand{\data}[1]{\texttt{#1}\xspace}
\newcommand{\mlone}{\data{ml1m}}
\newcommand{\mlten}{\data{ml10m}}
\newcommand{\net}{\data{netflix}}
\newcommand{\wiki}{\data{wiki-simple}}

\newcommand{\method}[1]{{\sf #1}\xspace}
\newcommand{\fullss}{\method{GD-diag}}
\newcommand{\fullsogram}{\method{SOGram-diag}}
\newcommand{\fullsg}{\method{Sampling-diag}}
\newcommand{\gd}{\method{GD}}
\newcommand{\mbss}{\method{Sampling}}
\newcommand{\mbsogram}{\method{SOGram}}
\newcommand{\gn}{\method{Newton}}

\newcommand{\costofforward}[1]{F(#1)}

\def\Figref#1{Figure~\ref{#1}}

\def\Tabref#1{Table~\ref{#1}}
\def\Secref#1{Section~\ref{#1}}
\def\eqref#1{(\ref{#1})}
\def\Algref#1{Algorithm~\ref{#1}}

\def\1{\bm{1}}

\def\vtheta{{\bm{\theta}}}
\def\va{{\bm{a}}}
\def\vb{{\bm{b}}}

\def\vd{{\bm{d}}}

\def\vh{{\bm{h}}}

\def\vp{{\bm{p}}}
\def\vq{{\bm{q}}}
\def\vr{{\bm{r}}}
\def\vs{{\bm{s}}}

\def\vu{{\bm{u}}}
\def\vv{{\bm{v}}}
\def\vw{{\bm{w}}}

\def\mA{{\bm{A}}}
\def\mB{{\bm{B}}}

\def\mG{{\bm{G}}}
\def\mH{{\bm{H}}}
\def\mI{{\bm{I}}}

\def\mM{{\bm{M}}}

\def\mP{{\bm{P}}}
\def\mQ{{\bm{Q}}}
\def\mR{{\bm{R}}}

\def\mW{{\bm{W}}}
\def\mX{{\bm{X}}}
\def\mY{{\bm{Y}}}
\def\mZ{{\bm{Z}}}

\DeclareMathAlphabet{\mathsfit}{\encodingdefault}{\sfdefault}{m}{sl}
\SetMathAlphabet{\mathsfit}{bold}{\encodingdefault}{\sfdefault}{bx}{n}
\newcommand{\tens}[1]{\bm{\mathsfit{#1}}}

\def\tJ{{\tens{J}}}

\def\gO{{\mathcal{O}}}

\def\sB{{\mathbb{B}}}

\def\sO{{\mathbb{O}}}

\def\sR{{\mathbb{R}}}

\def\sU{{\mathbb{U}}}
\def\sV{{\mathbb{V}}}

\newcommand{\reg}{\lambda}

\usepackage{subcaption}
\usepackage{enumitem}

\title{Efficient Optimization Methods for Extreme Similarity Learning with Nonlinear Embeddings}

\myexternaldocument{supplement}

\copyrightyear{2021}
\acmYear{2021}
\setcopyright{acmcopyright}\acmConference[KDD '21]{Proceedings of the 27th ACM SIGKDD Conference on Knowledge Discovery and Data Mining}{August 14--18, 2021}{Virtual Event, Singapore}
\acmBooktitle{Proceedings of the 27th ACM SIGKDD Conference on Knowledge Discovery and Data Mining (KDD '21), August 14--18, 2021, Virtual Event, Singapore}
\acmPrice{15.00}
\acmDOI{10.1145/3447548.3467363}
\acmISBN{978-1-4503-8332-5/21/08}

\begin{document}
%\thinmuskip=0mu\medmuskip=0mu\thickmuskip=0mu
%\abovedisplayskip=2.1pt
%\belowdisplayskip=2.1pt
%\abovedisplayshortskip=0.2pt
%\belowdisplayshortskip=0.2pt

\fancyhead{}

\author{Bowen Yuan}
\affiliation{%
  \institution{National Taiwan University}
  \city{}
  \country{}
}
\email{f03944049@csie.ntu.edu.tw}

\author{Yu-Sheng Li}
\affiliation{%
  \institution{National Taiwan University}
   \city{}
  \country{}
}
\email{r07922087@csie.ntu.edu.tw}

\author{Pengrui Quan}
\affiliation{%
  \institution{University of California, Los Angeles}
  \city{}
  \state{}
  \country{}
}
\email{prquan@g.ucla.edu}

\author{Chih-Jen Lin}
\affiliation{%
  \institution{National Taiwan University}
    \city{}
  \country{}
}
\email{cjlin@csie.ntu.edu.tw}

\begin{abstract}
We study the problem of learning similarity by using nonlinear embedding models (e.g., neural networks) from all possible pairs. This problem is well-known for its difficulty of training with the extreme number of pairs. For the special case of using linear embeddings, many studies have addressed this issue of handling all pairs by considering certain loss functions and developing efficient optimization algorithms. This paper aims to extend results for general nonlinear embeddings. First, we finish detailed derivations and provide clean formulations for efficiently calculating some building blocks of optimization algorithms such as function, gradient evaluation, and Hessian-vector product. The result enables the use of many optimization methods for extreme similarity learning with nonlinear embeddings. Second, we study some optimization methods in detail. Due to the use of nonlinear embeddings, implementation issues different from linear cases are addressed. In the end, some methods are shown to be highly efficient for extreme similarity learning with nonlinear embeddings. 
\end{abstract}



\maketitle

\SetKwInput{KwInit}{Init}
\SetKw{Break}{break}
\SetAlgoNoLine

\section{Introduction}
\label{sec:intro}
Many applications can be cast in the problem of learning similarity between a pair of two entities referred to as the left and the right entities respectively.
For example, in recommender systems, the similarity of a user-item pair indicates the preference of the user on the item.
In search engines, the similarity of a query-document pair can be used as the relevance between the query and the document.
In multi-label classifications, for any instance-label pair with high similarity, the instance can be categorized to the label.

A popular approach for similarity learning is to train an embedding model as the representation of each entity in the pair, such that any pair with high similarity are mapped to two close vectors in the embedding space, and vice versa. For the choice of the embedding model, some recent works \citep{KZ19a-short} report the superiority of nonlinear over conventional linear ones. 
A typical example of applying nonlinear embedding models is the two-tower structure illustrated in \Figref{fig:twin}, where a multi-layer neural network serves as the embedding model of each entity. Some successful uses in real-world applications have been reported \citep{PSH13a-short, AME15a-short, YY18a-short, XY19a-short, JTH20a-short}.
%TODO update citations 

While many works only consider observed pairs for similarity learning, more and more works argue that better performance can be achieved by considering all possible pairs. 
For example, recommender systems with implicit feedback face a one-class scenario, where all observed pairs are labeled as similar while the dissimilar ones are missing. 
To achieve better similarity learning, a widely-used setting \citep{RP09a-short, HFY14b-short, HFY17a-short} is to include all unobserved pairs as dissimilar ones. Another example is counterfactual learning \citep{BY19b-short, XW19a-short}, where because the observed pairs carry the selection bias caused by confounders, an effective way to eliminate the bias is to additionally consider the unobserved pairs by imputing their labels. 

However, for many real-world scenarios, both the number $m$ of left entities and the number $n$ of right entities can be tremendous.
The learning procedure is challenging 
as a prohibitive $\bbO{mn}$ cost occurs by directly applying any optimization algorithm.
To avoid the $\bbO{mn}$ complexity, one can subsample unobserved pairs, but it has been shown that such subsampled settings are inferior to the non-subsampled setting \citep{RP08a-short, HFY16a-short}.
We refer to the problem of the similarity learning from extremely large $mn$ pairs as {\em extreme similarity learning}.

\newcommand\tower[5]{
\draw[rounded corners=5pt] (#1+1,4) rectangle ++(#2-2,0.6) node[pos=.5] {#3};
\draw[rounded corners=5pt] (#1+1,1.4) rectangle ++(#2-2,0.6) node[pos=.5] {#4};
\node[minimum size=15pt,inner sep=0pt] (dot) at (#1+2,3) {#5};
\node [trapezium,trapezium angle=80,inner xsep=0pt,minimum height=1.5cm,text width=.3cm,draw] at (#1+#2/2, 3) {};
}

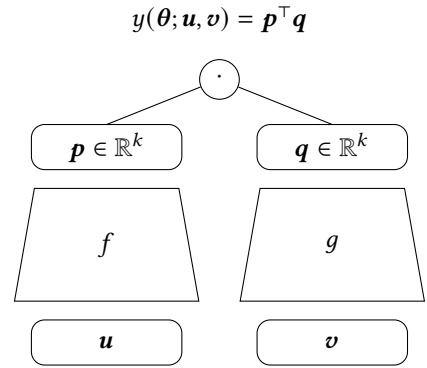
\begin{figure}
\centering
\scalebox{1.0}{
\begin{tikzpicture}[]
\large
\tower{1}{4}{$\vp \in \sR^k$}{$\vu$}{$f$}
\tower{4}{4}{$\vq \in \sR^k$}{$\vv$}{$g$}
\node[minimum size=15pt,inner sep=0pt] (dot) at (4.5, 6) {$y(\vtheta; \vu, \vv) = \vp^\top \vq$};
\node[circle, draw, minimum size=15pt,inner sep=0pt] (dot) at (4.5, 5.2) {$\cdot$};
\draw (3, 4.6) -- (dot);
\draw (6, 4.6) -- (dot);
\end{tikzpicture}}
\caption{An illustration of the two-tower structure, adopted from \cite{SR20a-short}.}
\Description{An illustration of the two-tower structure; modified from \cite{SR20a-short}.}
\label{fig:twin}
\end{figure}

%For past works of extreme similarity learning, most of them only consider linear embeddings (e.g., matrix factorization in \citep{RP09a, XH16b, HFY16a}) , while few studies non-linear embeddings. 
%A review will be given in Section \ref{sec:rw}, where through describing how these works take advantage of a key feature of linear embeddings, we point out challenging issues of extreme similarity learning with nonlinear embeddings. 

To tackle the high $\gO(mn)$ complexity, the aforementioned works that can handle all pairs consider a loss function with a certain structure for the unobserved pairs.
Then they are able to replace the $\bbO{mn}$ complexity with a much smaller $\bbO{m+n}$ one.  In particular, the function value, gradient, or other information can be calculated in $\bbO{m+n}$ cost, so various types of optimization methods can be considered. Most exiting works consider linear embeddings (e.g., matrix factorization in \citep{RP09a-short, XH16b-short, HFY16a-short}), where the optimization problem is often in a multi-block convex form. Thus many consider a block-wise setting to sequentially minimize convex sub-problems. The needed gradient or other information on each block of variables can also be calculated in  $\bbO{m+n}$.

If general nonlinear embeddings are considered, so far few works have studied the optimization algorithm. This work aims to fill the gap with the following main contributions.
\begin{itemize}[leftmargin=*]
  \item To calculate function value, gradient, or other information in $\bbO{m+n}$ cost, the extension from cases in linear embeddings seems to be possible, but detailed derivations have not been available. We finish tedious calculations and provide clean formulations. This result enables the use of many optimization methods for extreme similarity learning with nonlinear embeddings.
  \item We then study some optimization methods in detail. Due to the use of nonlinear embeddings, some implementation issues different from linear cases are addressed. In the end, some methods are shown to be highly efficient for extreme similarity learning with nonlinear embeddings.
\end{itemize}

The paper is organized as follows. A review on extreme similarity learning is in \Secref{sec:prel}. We derive efficient computation for some important components in \Secref{sec:fw}. In Section \ref{sec:eom}, we demonstrate that many optimization algorithms can be used for extreme similarity learning with nonlinear embeddings. Section \ref{sec:imp} describes some implementation issues.
Finally, we present experiments  on large-scale data sets in \Secref{sec:exp}, and conclude in \Secref{sec:con}. Table \ref{tab:notations} gives main notations in this paper. The supplementary materials and data/code for experiments are available at \url{https://www.csie.ntu.edu.tw/~cjlin/papers/similarity_learning/}.

\begin{table}[t]
\caption{Main notation}
\label{tab:notations}
\centering
  \addtolength{\tabcolsep}{-3pt}

                \begin{tabular}{l|l}
                        \toprule
                        Notation & Description \\
                        \midrule
                        $(i, j)$ & (left entity $i$, right entity $j$) pair\\
                        $m, n$ & numbers of  left entities and right entities\\
                        $\vu, \vv $ & feature vectors of a left entity and a right entity\\
                        $\obs$ & set of observed pairs \\
%                        $D = D_u + D_v$ & total number of variables \\
                        $y(\cdot),\vtheta \in \sR^{D_u+D_v}$ & a similarity function and its variables \\
%                        $\vtheta$ & parameters of similarity function\\
                        $L(\vtheta)$ & objective function\\
                        $\mR$ & observed incomplete similarity matrix\\
                        $\hat{\mY}$ & similarity matrix predicted by $y(\cdot)$\\
                        $\tilde{\mY}$ &imputed similarity matrix (for unobserved pairs)\\
                        $f(\cdot), g(\cdot)$ & embedding models of left and right entities\\
                        $\vp \in \sR^k, \vq \in \sR^k$ & embedding vectors of left and right entities\\
%                        $k$ & embedding size\\
                      	$\costofforward{f}$, $\costofforward{g}$& cost of operations related to $f(\cdot)$, and $g(\cdot)$ \\
                        \bottomrule
                \end{tabular}
\end{table}
\section{Extreme Similarity Learning}
We review extreme similarity learning problems.
\label{sec:prel}

\subsection{Problem Formulation}
%Let $m$ and $n$ be respectively the number of left and right entities. 
Many real-world applications can be modeled as a learning problem with an incomplete similarity matrix $\mR$ of $m$ left entities and $n$ right entities. The set of observed pairs is denoted as $\obs \subseteq \{1, \dots, m\} \times \{1, \dots, n\}$ where typically $|\obs| \ll mn$.
Only $R_{ij}, \forall (i, j) \in \obs$ are revealed.
%Let $\obs \subseteq \{1, \dots, m\} \times \{1, \dots, n\}$ be the set of observed pairs, i.e. $\obs = \{(i, j) : Y_{ij} = 1\}$. Usually $|\obs| \ll mn$.
Besides $\mR$, we assume that side features $\sU = \{\dots,\vu_i, \dots\}$ of left entities and $\sV = \{\dots,\vv_j, \dots\}$ of right entities are available.
Our goal is to find a similarity function $y: \sR^{D} \to \sR$,
 \begin{equation}
  \hYij = y(\vtheta; \vu_i, \vv_j),
\label{eq:last_inner}
\end{equation}
where $\vtheta = \begin{bmatrix} \vtheta^u \\ \vtheta^v \end{bmatrix}$ is the vectorized variables of $y$ with $\vtheta^u \in \sR^{D_u}, \vtheta^v \in \sR^{D_v}$, and $D = D_u + D_v$.
For the similarity function, we consider the following dot product similarity that has been justified in a recent comparative study of similarity functions in recommender systems \citep{SR20a-short}
\begin{displaymath}
  y(\vtheta; \vu_i, \vv_j) = f(\vtheta^u; \vu_i)^\top g(\vtheta^v; \vv_j),
\end{displaymath}
where $f:\sR^{D_u}\to \sR^k$ and $g:\sR^{D_v}\to \sR^k$ are two embedding models that learn representations of left and right entities, respectively; see \Figref{fig:twin}. In this work, we focus on nonlinear embedding models.
For convenience, we  abbreviate $f(\vtheta^u; \vu_i)$ and $g(\vtheta^v; \vv_j)$ to $\Fi$ and $\Fj$, respectively.
We also write the embedding vectors as
\begin{equation} \label{eq:piqj}
\Zi = \Fi, \Zj = \Fj
\end{equation}
so that 
\begin{equation}
  \hYij = \Zi^\top \Zj = \Fi^\top \Fj.
\label{eq:rush}
\end{equation}

By considering all $mn$ pairs, we learn $\vtheta$ through solving
\begin{equation}
  \min\nolimits_{\vtheta} \quad \La,
\label{eq:prob}
\end{equation}
where the objective function is
\begin{equation}
 \La = \csum_{i=1}^m\csum_{j=1}^n \lij(Y_{ij}, \hYij) + \lambda \regularizer(\vtheta),
 \label{eq:general_funval}
\end{equation}
$\lij$ is an entry-wise twice-differentiable loss function convex in $\hYij$, $\regularizer(\vtheta)$ is a twice-differentiable and strongly convex regularizer to avoid overfitting, and $\lambda > 0$ is the corresponding parameter.
We assume $\regularizer$ is simple so that in all complexity analyses we ignore the costs related to $\regularizer(\vtheta), \nabla \regularizer(\vtheta)$, and $\nabla^2 \regularizer(\vtheta)$.
In \eqref{eq:general_funval},
we further assume
\begin{equation} 
%Y_{ij}=R_{ij}, \forall (i, j) \in \obs,\quad \text{otherwise }Y_{ij} = \tilde{Y}_{ij},
Y_{ij}=
 \begin{cases}
	R_{ij} & (i, j) \in \obs,\\
	\tilde{Y}_{ij} & (i, j) \notin \obs,
\end{cases}
\end{equation}
where for any unobserved pair $(i, j) \notin \obs$ we impute $\tilde{Y}_{ij}$ as an artificial label of the similarity. 
For instance, in recommender systems with implicit feedback, by treating unobserved pairs as negative, usually $\tilde{Y}_{ij} = 0$ or $-1$ is considered. 

Clearly \eqref{eq:general_funval} involves a summation of $mn$ terms, so the cost of directly applying an optimization algorithm is prohibitively proportional to $\bbO{mn}$.

%We use $\costofforward{f}$ and $\costofforward{g}$ respectively to represent the cost of operations related to the two nonlinear embedding models.
%If neural networks are considered, and let $\vd^u, \vd^v$ be some vectors, then the costs include
%\begin{itemize}[leftmargin=*]
%  \item the forward computation of $\Fi, \forall i$ and $\Fj, \forall j$
%  \item the transposed Jacobian-vector products $\tJfi \vd^u, \forall i$ and $\tJgj \vd^v, \forall j$
%  \item the Jacobian-vector products $\Jfi \vd^u, \forall i$ and $\Jgj \vd^v, \forall j$
%\end{itemize}
%In \Secref{sec:imp} for implementation details, we will show that the complexity of these operations differs only up to a constant factor.
%Thus we can use the same notation to broadly cover them.

\subsection{Linear Embedding Models}
\label{sec:rw}
As mentioned in Section \ref{sec:intro}, most existing works of extreme similarity learning focus on linear embedding models. 
To avoid any $\bbO{mn}$ cost occurs in solving the optimization problem, they consider a special kind of loss functions
\begin{equation}
  \lij(\Yij, \hYij) = 
\begin{cases}
	\ell(\Yij, \hYij) & (i, j) \in \obs,\\
	\half \nw \wi\wj (\tYij - \hYij)^2 & (i, j) \notin \obs,
\end{cases}
\label{eq:loss}
\end{equation}
where $\ell(a, b)$ is any non-negative and twice-differentiable function convex in $b$,
and $\nw$ is a parameter for balancing the two kinds of losses. Two vectors
$\va \in \sR^{m}$, and $\vb \in \sR^{n}$ are chosen so that $\wi\wj$ is a cost associated with an unobserved pair.
Besides, for the imputed label $\tYij$, it is required \citep{WK19a-short, BY19b-short} that
\begin{equation}
  \tYij = \tZi^\top\tZj,
\label{eq:tYij}
\end{equation}
where $\tZi, \tZj$ are vectors output by the prior imputation model, and are fixed throughout training.
For easy analysis, we assume that $\tZi, \tZj \in \sR^k$ have the same length as $\Zi, \Zj$. 
With the loss in \eqref{eq:loss}, past works were able to reduce the $\bbO{mn}$ cost to $\bbO{m+n}$. The main reason is because derived computations
on all $i$ and all $j$ can be obtained by using values solely related to
$i$ and $j$, respectively. A conceptual illustration is in the following equation.
\begin{equation}
  \csum_{i=1}^m\csum_{j=1}^n(\cdots)=\csum_{i=1}^m(\cdots) {\text{ } \hspace{0.2cm} \times } \hspace{0.2cm} \csum_{j=1}^n(\cdots).
\label{eq:mn_trick}
\end{equation}
The mainline of most past works is to incorporate more linear models into extreme similarity learning, which can range from simple matrix factorization \citep{RP09a-short, XH16b-short, HFY16a-short} to complex models, e.g., matrix factorization with side information and (field-aware) factorization machines \citep{IB17a-short, HFY17a-short, BY19a-short}.

%\begin{algorithm}[t]
%    \caption{A general optimization framework for solving \eqref{eq:general_funval} with non-linear embedding models}
%    \label{alg:framework}
%  \DontPrintSemicolon
%	\KwInit{
%    Draw $\vtheta$ randomly
%  }
%  \Repeat{\text{stopping condition is satisfied}}{
%    Compute (approximated) gradient information $\grad$ \\
%    Obtain direction $\vs$ with  $\grad$\\
%    Find a step size $\delta$\\
%    Update $\vtheta \gets \vtheta + \delta \vs$
%  }
%\end{algorithm}

\section{Information Needed in Optimization Algorithms}
\label{sec:fw}
 Most optimization methods involve a procedure that repeatedly calculates the function value, the gradient, and/or other information. For example, a gradient descent method must calculate the negative gradient as the direction for an update. Crucially, from past developments for linear embeddings, the $\bbO{mn}$ cost mainly occurs in these building blocks of optimization algorithms. The aim of this section is to derive $\bbO{m+n}$ computation for these important components when nonlinear embeddings are applied. 

\subsection{From Linear to Nonlinear Embedding Models}
\label{sec:from_linear}
We begin by discussing the differences between using linear and nonlinear embeddings. If linear embeddings are used, a key feature is that \eqref{eq:prob} becomes a multi-block convex problem. That is, \eqref{eq:prob} is reduced to a convex sub-problem when embedding variables of one side are fixed. Then existing works apply alternating least squares \citep{RP09a-short}, (block) coordinate descent \citep{XH16b-short, HFY16a-short,IB17a-short, BY19a-short}, and alternating Newton method \citep{HFY17a-short, BY19b-short} to sequentially solve each sub-problem by convex optimization techniques, which often iteratively calculate the following information.
%the function value, the gradient, and/or the Hessian-vector product.
\begin{itemize}[leftmargin=*]
  \item Function value,
  \item Gradient, and/or
  \item Hessian-vector product
\end{itemize}
Thus a focus was on deriving $\bbO{m+n}$ operations for them.

If nonlinear embeddings are considered, in general \eqref{eq:prob} is not a block-convex problem. Thus a block-wise setting may not possess advantages so that optimization algorithms updating all variables together is a more natural choice. In any case, if we can obtain formulations for calculating the function value or the gradient over all variables, then the information over a block of variables is often in a simplified form. For example, the gradient over the block $\vtheta^u$ is a sub-vector of the whole gradient over $\vtheta = \begin{bmatrix} \vtheta^u \\ \vtheta^v \end{bmatrix}$. Therefore, in the rest of this section, we check the general scenario of considering all variables.

%In this section, after address the training efficiency issue of some key computations in this framework, we derive several efficient methods for extreme similarity learning with nonlinear embeddings. 

%\begin{algorithm}[t]
%    \caption{A general optimization framework for solving \eqref{eq:general_funval} with non-linear embedding models}
%    \label{alg:framework}
%  \DontPrintSemicolon
%	\KwInit{
%    Draw $\vtheta$ randomly
%  }
%  \Repeat{\text{stopping condition is satisfied}}{
%    Compute (approximated) gradient information $\grad$ \\
%    Obtain direction $\vs$ with  $\grad$\\
%    Find a step size $\delta$\\
%    Update $\vtheta \gets \vtheta + \delta \vs$
%  }
%\end{algorithm}
\subsection{Evaluation of the Objective Value}
\label{sec:funval}
The infeasible $\bbO{mn}$ cost occurs if we 
directly compute $\La$ in \eqref{eq:general_funval}. To handle this, our idea is to follow past works of linear embeddings to consider \eqref{eq:loss} as the loss function. 

By applying \eqref{eq:loss}, \eqref{eq:general_funval} is equivalent to
\begin{align}
\La =& \ssum \ell(\Yij, \hYij)+\sum_{(i,j)\notin \obs} \half\nw\wi\wj (\tYij-\hYij)^2+\lambda \regularizer(\vtheta)\nonumber \\
=&\underbrace{\ssum \ell^+_{ij}(\Yij, \hYij)}_{\Lp} + \nw \underbrace{\dbsum \ell^-_{ij}(\Yij, \hYij)}_{\Ln}+ \lambda \regularizer(\vtheta),\label{eq:obj}
\end{align}
where
\begin{align}
  \ell^+_{ij}(\Yij, \hYij) &= \ell(\Yij, \hYij)- \half \nw \wi\wj (\tYij-\hYij)^2, \nonumber \\
  \ell^-_{ij}(\Yij, \hYij) &= \frac{1}{2}\wi\wj(\tYij-\hYij)^2. \label{eq:lij}
\end{align}
The first term $\Lp$ involves the summation of $|\obs|$ values, so with $|\obs| \ll mn$ the bottleneck is on the second term $\Ln$.

Let $\costofforward{f}$ and $\costofforward{g}$ respectively  represent the cost of operations related to the two nonlinear embedding models,\footnote{If neural networks are considered, operations such as forward or backward computation are considered. See more discussion in Section \ref{sec:imp}} and
\begin{equation} \label{eq:PQ}
\mP = \begin{bmatrix}
  \vp_1^\top \\
  \vdots \\
  \vp_m^\top
\end{bmatrix},
\mQ = \begin{bmatrix}
  \vq_1^\top \\
  \vdots \\
  \vq_n^\top
\end{bmatrix},
\tilde \mP = \begin{bmatrix}
  \tilde \vp_1^\top \\
  \vdots \\
  \tilde \vp_m^\top
\end{bmatrix},
\tilde \mQ = \begin{bmatrix}
  \tilde \vq_1^\top \\
  \vdots \\
  \tilde \vq_n^\top
\end{bmatrix},
\end{equation}
where $\tilde \mP$ and $\tilde \mQ$ are constant matrices, but $\mP$ and $\mQ$ depend on $\vtheta$.
To compute $\Lp$ in \eqref{eq:obj}, we first compute $\mP$ and $\mQ$ in $\bbO{m \costofforward{f}+n \costofforward{g}}$ time and store them in $\bbO{(m+n)k}$ space. Then for each $(i, j) \in \obs$, from \eqref{eq:rush} and \eqref{eq:tYij},
we compute $\hYij=\Zi^\top\Zj$ and $\tYij = \tZi^\top \tZj$ in $\bbO{k}$ time. As a result, we can compute the entire $\Lp$ in $\bbO{m \costofforward{f}+n \costofforward{g} + |\obs|k}$ time.

For $\Ln$, we follow the idea in \eqref{eq:mn_trick} to derive
\begin{equation}\label{eq:Ln_brief}
  \Ln = \half \fip{\tPc}{\tQc}-\fip{\hPc}{\hQc}+ \half \fip{\Pc}{\Qc},
\end{equation}
where details are left in Appendix \ref{sec:D_fun_val}. Later for gradient calculation we show details as a demonstration of using \eqref{eq:mn_trick}.

In \eqref{eq:Ln_brief}, $\fip{\cdot}{\cdot}$ is a Frobenius inner product between two matrices,
\begin{equation}
\begin{aligned}
  &\Pc = \mP^\top \Wu \mP,\hPc=\tilde{\mP}^\top \Wu \mP,\tPc =\tilde{\mP}^\top\Wu \tilde{\mP},\\  &\Qc = \mQ^\top \Wv \mQ,\hQc =\tilde{\mQ}^\top \Wv \mQ, \text{ and } \tQc=\tilde{\mQ}^\top \Wv \tilde{\mQ},
\end{aligned}
\label{eq:gramian}
\end{equation}
where $\Wu = \diag(\va), \Wv = \diag(\vb)$ are two diagonal matrices.
As $\mP \in \sR^{m\times k}$ and $\mQ \in \sR^{n \times k}$ have been pre-computed and cached in memory during computing $\Lp$, the matrices in \eqref{eq:gramian} can be computed in $\bbO{(m+n)k^2}$ time and cached in $\bbO{k^2}$ space. Then the Frobenius inner products between these matrices cost only $\bbO{k^2}$ time. 

From complexities of $\Lp$ and $\Ln$, the overall cost of 
evaluating $\La$ is 
\begin{equation}
  \bbO{|\obs|k + (m+n)k^2+ m \costofforward{f}+n \costofforward{g}}.
\label{eq:com_fun_val}
\end{equation}

\subsection{Computation of Gradient Information} \label{sec:grad}
For $\La$ defined in \eqref{eq:general_funval}, the gradient is
\begin{equation}
  \grad = \dbsum \tJyij \parti{\lij}{\hYij} + \lambda \nabla \regularizer(\vtheta).
 \label{eq:general_grad}
\end{equation}
Analogous to \eqref{eq:general_funval}, it is impractical to directly compute $\grad$ with infeasible $\gO(mn)$ costs.
For derivatives with respect to a vector, we let $\xJyij$ be a row vector, and later for operations such as $\xJfi$, we let it be a $k \times D_u$ matrix. In all other situations, a vector such as $\grad$ is a column vector.

From \eqref{eq:loss} and \eqref{eq:obj}, $\grad$ in \eqref{eq:general_grad} is equivalent to  
\begin{equation*}
\grad = \underbrace{\ssum \tJyij \parti{\lij^+}{\hYij}}_{\pgrad} + \nw \underbrace{\dbsum \tJyij \parti{\lij^-}{\hYij}}_{\ngrad} + \lambda \nabla \regularizer(\vtheta).
\end{equation*}
To calculate $\pgrad$, 
let $\mX \in \sR^{m\times n}$ be a sparse matrix with
\begin{equation}
  X_{ij} = \begin{cases}
    \parti{\lij^+}{\hYij} & (i, j) \in \obs, \\
    0 & \text{otherwise}.
  \end{cases}
\label{eq:Xij}
\end{equation}
From \eqref{eq:piqj} and \eqref{eq:rush},
\begin{equation}
  \Jyij = \Big[ \Jyf \Jfi, \Jyg \Jgj \Big]
  % = \Big[ \Fj^\top \Jfi, \Fi^\top \Jgj \Big]
  = \Big[ \Zj^\top \Jfi, \Zi^\top \Jgj \Big],
\label{eq:tJyij}
\end{equation}
so with \eqref{eq:Xij} we have
\begin{equation}
	\pgrad = \dbsum \Big(\tJyij X_{ij}\Big) =
  \begin{bmatrix}
    \dbsum \Jfi^\top \Zj X_{ij} \\
    \dbsum \Jgj^\top \Zi X_{ij}
  \end{bmatrix}.
\label{eq:pgrad}
\end{equation}
Let $\tJf \in \sR^{m \times D_u \times k}, \tJg \in \sR^{n \times D_v \times k}$ be tensors with 
\begin{equation}
 \text{the $i$-th slice $\tJf_{i, :, :} = \tJfi$ and the $j$-th slice
$\tJg_{j, :, :} = \tJgj$}, 
\label{eq:tensor_slice}
\end{equation}
respectively. 
%For tensor $\tJ \in \sR^{m \times D \times k}$, the notation can be extended similarly so that $\tJ_{i, :, :} \in \sR^{D \times k}$ represents the $i$-th slice of $\tJ$.
The top part of \eqref{eq:pgrad} can be computed by
\begin{align}
  \dbsum \tJfi \Zj X_{ij}
  =&\msum \tJfi \Big( \nsum X_{ij} \Zj^\top \Big)^\top \nonumber\\
  =&\msum \tJfi \Big( \mX_{i, :} \mQ \Big)^\top 
  =\langle \tJf, \mX \mQ \rangle, \label{eq:lp_grad}
\end{align}
where \eqref{eq:lp_grad} is from \eqref{eq:PQ} and the following definition of the tensor-matrix inner product 
\begin{equation} \label{eq:ten-mat}
   \langle \tJ, \mM \rangle = \msum \tJ_{i, :, :} (\mM_{i, :})^\top.
\end{equation}
As the computation of the bottom part of \eqref{eq:pgrad} is similar, we omit the details. Then \eqref{eq:pgrad} can be written as 
\begin{equation}
	\pgrad = \begin{bmatrix}
    \langle \tJf, \mX \mQ \rangle \\
    \langle \tJg, \mX^\top \mP \rangle
  \end{bmatrix},
\label{eq:pgrad_final}
\end{equation}
where $\mP$ and $\mQ$ are first computed in $\bbO{m \costofforward{f}+n \costofforward{g}}$ time and stored in $\bbO{(m+n)k}$ space.
Then we compute $\mX \mQ$ and  $\mX^\top \mP$ in $\bbO{|\obs|k}$ time and $\bbO{(m+n)k}$ space.

To calculate $\ngrad$, from \eqref{eq:rush} and \eqref{eq:lij} we have
  $$\parti{\lij^-}{\hYij} = \wi\wj(\Zi^\top\Zj-\tZi^\top \tZj).$$
This and \eqref{eq:tJyij} imply that
\begin{align}
  &\ngrad=\dbsum \begin{bmatrix} \tJfi\Zj \\ \tJgj\Zi \end{bmatrix} \wi\wj (\Zi^\top\Zj-\tZi^\top \tZj) \label{eq:ln_grad_init} \\
  =&\begin{bmatrix}
    \msum\tJfi \nsum \Zj \wi\wj (\Zj^\top\Zi-\tZj^\top \tZi) \\
    \nsum\tJgj \msum \Zi \wi\wj (\Zi^\top\Zj-\tZi^\top \tZj)
  \end{bmatrix} \nonumber \\
  =&\begin{bmatrix}
    \msum\tJfi\Big((\nsum\wj\Zj\Zj^\top)\wi\Zi - (\nsum\wj\Zj\tZj^\top)\wi\tZi\Big) \\
    \nsum\tJgj\Big((\msum\wi\Zi\Zi^\top)\wj\Zj - (\msum\wi\Zi\tZi^\top)\wj\tZj\Big)
  \end{bmatrix}, \nonumber 
\end{align}
where the last two equalities follow from the idea in \eqref{eq:mn_trick}. From \eqref{eq:gramian},
\begin{equation}
\begin{aligned}
  &\Pc = \textstyle{\msum} \wi\Zi\Zi^\top, \Qc = \nsum \wj\Zj\Zj^\top, \\ &\hPc = \textstyle{\msum} \wi\tZi\Zi^\top, \text{ and } \hQc = \nsum \wj\tZj\Zj^\top,
\end{aligned}
\label{eq:gramian_detail}
\end{equation}
so we can compute $\ngrad$ by
\begin{align}
  \ngrad=&\begin{bmatrix}
    \msum\tJfi(\Qc\wi\Zi - \hQc^\top\wi\tZi) \\
    \nsum\tJgj(\Pc\wj\Zj - \hPc^\top\wj\tZj)
  \end{bmatrix}  \nonumber\\
  =&\begin{bmatrix}
    \langle \tJf, \Wu \mP \Qc - \Wu \tilde \mP \hQc \rangle \\
    \langle \tJg, \Wv \mQ \Pc - \Wv \tilde \mQ \hPc \rangle
  \end{bmatrix}, \label{eq:ln_grad}
\end{align}
where \eqref{eq:ln_grad} follows from \eqref{eq:PQ}, \eqref{eq:tensor_slice}, and \eqref{eq:ten-mat}.
As $\mP$ and $\mQ$ have been pre-computed and cached in memory during computing $\pgrad$, all matrices in \eqref{eq:gramian} 
can be computed in $\bbO{(m+n)k^2}$ time and cached in $\bbO{k^2}$ space. 
Then the right hand sides of $\langle\cdot,\cdot\rangle$ in \eqref{eq:ln_grad} can be computed in $\bbO{(m+n)k^2}$ time as $\Wu$ and $\Wv$ are diagonal.

By combining \eqref{eq:pgrad_final} and \eqref{eq:ln_grad}, the entire $\grad$ can be efficiently computed by
\begin{equation}
  \grad = \begin{bmatrix}
    \langle \tJf, \mX \mQ + \nw(\Wu \mP \Qc - \Wu \tilde \mP \hQc) \rangle \\
    \langle \tJg, \mX^\top \mP + \nw(\Wv \mQ \Pc - \Wv \tilde \mQ \hPc) \rangle
  \end{bmatrix} + \lambda \nabla \regularizer(\vtheta),
\label{eq:final_grad}
\end{equation}
which has a time complexity of
\begin{equation}
  \bbO{|\obs|k + (m+n)k^2+ m \costofforward{f}+n \costofforward{g}}.
\label{eq:grad_comp}
\end{equation}
The $m \costofforward{f}+n \costofforward{g}$ terms come from calculating $\mP$ and $\mQ$ in \eqref{eq:pgrad_final} and from
$\xJfi, \forall i$, $\xJgj, \forall j$ in \eqref{eq:tensor_slice}.
The procedure of evaluating $\grad$ is summarized in \Algref{alg:grad-detailed}.

From \eqref{eq:com_fun_val} and \eqref{eq:grad_comp}, the cost of one function evaluation is similar to that of one gradient evaluation.

\begin{algorithm}[t]
    \caption{Gradient evaluation: $\grad$}
    \label{alg:grad-detailed}
  \DontPrintSemicolon
	\KwIn{$\vtheta, \obs, \tilde{\mP}, \tilde{\mQ},\Wu, \Wv$.}
	 $\text{Calculate } \mP, \mQ, \Pc, \Qc, \hPc, \hQc.$\footnotemark\\
    $X_{ij} \gets \parti{\lij^+}{\hYij}, \forall (i, j)\in \obs$ \\
    $\mX_q \gets \mX\mQ$, $\mX_p \gets \mX^\top \mP$\\
    $\grad \gets \begin{bmatrix} \langle \tJf, \mX_q + \nw(\Wu \mP \Qc - \Wu \tilde \mP \hQc) \rangle \\ \langle \tJg, \mX_p + \nw(\Wv \mQ \Pc - \Wv \tilde \mQ \hPc)\rangle \end{bmatrix} + \lambda \nabla \regularizer(\vtheta).$\\
    \KwOut{$\grad$}
\end{algorithm}
\footnotetext{If for the same $\vtheta$, $\La$ is calculated before $\grad$, then these matrices have been stored and are available as input.}

\subsection{Computation of Gauss-Newton Matrix-vector Products}
\label{sec:gnmvp}
For optimization methods using second-order information, commonly the product between Hessian and some vector $\vd$ is needed:
\begin{equation}
  \hess\vd.
\end{equation}
The Hessian of \eqref{eq:general_funval} is 
\begin{equation}
  \hess = \dbsum \tJyij \sparti{\lij}{\hYij} \Jyij + \dbsum \sparti{\hYij}{\vtheta} \parti{\lij}{\hYij} + \lambda \nabla^2 \regularizer(\vtheta).
 \label{eq:general_hess}
\end{equation}
Clearly, a direct calculation costs $\bbO{mn}$. However, before reducing the $\bbO{mn}$ cost to $\bbO{m+n}$, we must address an issue that $\hess$ is not positive definite. In past studies of linear embeddings (e.g., \cite{HFY17a-short, WSC18a-short, BY19a-short}) the block-convex $\La$ implies that for a strictly convex sub-problem, the Hessian is positive definite. Without such properties, here we need a positive definite  approximation of $\hess$. Following \cite{JM10a-short, CCW15a-short}, we remove the second term in \eqref{eq:general_hess} to have the following Gauss-Newton matrix \citep{NNS02a-short}
\begin{equation}
  \mG = \cdbsum \tJyij \sparti{\lij}{\hYij} \Jyij + \lambda \nabla^2 \regularizer(\vtheta),
\label{eq:general_G}
\end{equation}
which is positive definite from the convexity of $\lij$ in $\hYij$ and the strong convexity of $\regularizer(\vtheta)$. The matrix-vector product becomes
\begin{equation}
  \mG \vd = \cdbsum \tJyij \sparti{\lij}{\hYij} \Jyij \vd + \lambda \nabla^2 \regularizer(\vtheta) \vd.
\label{eq:general_Gd}
\end{equation}

From \eqref{eq:lij} we have 
\begin{equation}
  \sparti{\lij^-}{\hYij} = \wi\wj.
\label{eq:second_ij}
\end{equation}
Following \eqref{eq:obj}, we use \eqref{eq:second_ij} to re-write $\mG\vd$ in \eqref{eq:general_Gd} as
\begin{equation}
\begin{split}
  \underbrace{\ssum \tJyij \sparti{\lij^+}{\hYij} \Jyij \vd}_{\mG^+\vd}
  + \nw \underbrace{\dbsum \wi\wj \tJyij \Jyij \vd}_{\mG^-\vd} +\lambda \nabla^2 \regularizer(\vtheta) \vd.
\end{split}
\label{eq:Gd}
\end{equation}
Let $\vd = \begin{bmatrix} \vd^u \\ \vd^v \end{bmatrix}$ be some vector. By defining
\begin{equation}
  \bwi = \Jfi\vd^u \in \sR^{k}, \bhj = \Jgj\vd^v \in \sR^{k},
  \label{eq:Jvp}
\end{equation}
from \eqref{eq:tJyij}, we have
\begin{equation} \label{eq:Jd}
  \Jyij \vd = \Zj^\top\bwi + \Zi^\top\bhj.
\end{equation}

To compute $\mG^+\vd$ in \eqref{eq:Gd}, we first compute
$$\mW = \begin{bmatrix}
  \vw_1^\top\\ \vdots\\ \vw_m^\top
\end{bmatrix}\in \sR^{m \times k} \text{ and }
\mH = \begin{bmatrix}
  \vh_1^\top\\ \vdots \\\vh_n^\top
\end{bmatrix} \in \sR^{n \times k}$$
in $\bbO{m\costofforward{f} + n\costofforward{g}}$ time and store them in $\bbO{(m+n)k}$ space.
From \eqref{eq:Jd}, let $\mZ \in \sR^{m\times n}$ be a sparse matrix with
\begin{displaymath}
  Z_{ij} = \begin{cases}
    \sparti{\ell^+_{ij}}{\hYij}\Jyij\vd = \sparti{\lij^+}{\hYij}(\Zj^\top\bwi + \Zi^\top\bhj) & (i, j) \in \obs, \\
    0 & \text{otherwise},
  \end{cases}\end{displaymath}
which can be constructed in $\bbO{|\obs|k}$ time. 
Then from \eqref{eq:pgrad} and 
similar to the situation of computing \eqref{eq:lp_grad} and \eqref{eq:pgrad_final}, we have
\begin{equation}
  \mG^+\vd=\ssum \tJyij \sparti{\lij^+}{\hYij} \Jyij \vd = \dbsum \tJyij Z_{ij} =   \begin{bmatrix}
     \langle \tJf, \mZ \mQ \rangle \\
     \langle \tJg, \mZ^\top \mP \rangle
  \end{bmatrix},
\label{eq:lp_Gd}
\end{equation}
where $\mZ \mQ$ and $\mZ^\top \mP$ can be computed in $\bbO{|\obs|k}$ time and $\bbO{(m+n)k}$ space.

By using \eqref{eq:tJyij} and \eqref{eq:Jd}, $\mG^-\vd$ in \eqref{eq:Gd} can be computed by
\begin{align}
  \mG^-\vd
  =&\cdbsum \wi\wj \tJyij \Jyij \vd \nonumber \\
%  =&\dbsum \tJyij \wi\wj (\Zj^\top\bwi + \Zi^\top\bhj) \nonumber \\
  =&\cdbsum \begin{bmatrix} \tJfi\Zj \\ \tJgj\Zi \end{bmatrix} \wi\wj (\bwi^\top\Zj + \Zi^\top\bhj) \label{eq:ln_Gd0}\\
  =&\begin{bmatrix} \langle \tJf, \Wu \mW \Qc + \Wu \mP \mH_c \rangle \\ \langle \tJg, \Wv \mH \Pc + \Wv \mQ \mW_c\rangle \end{bmatrix}, \label{eq:ln_Gd}
\end{align}
where \eqref{eq:ln_Gd} is by a similar derivation from \eqref{eq:ln_grad_init} to \eqref{eq:ln_grad}, and by first computing the following $k \times k$ matrices in $\bbO{(m+n)k^2}$ time:
\begin{displaymath}
  \mW_c = \cmsum\wi\bwi\Zi^\top = \mW^\top \Wu \mP, \quad \mH_c = \cnsum\wj\bhj\Zj^\top = \mH^\top \Wv \mQ.
\end{displaymath}

Combining \eqref{eq:lp_Gd} and \eqref{eq:ln_Gd}, we can compute \eqref{eq:Gd} by
\begin{equation} \label{eq:final_Gd}
    \mG\vd = \begin{bmatrix} \langle \tJf, \mZ \mQ + \nw(\Wu \mW \Qc + \Wu \mP \mH_c) \rangle \\ \langle \tJg, \mZ^\top \mP + \nw(\Wv \mH \Pc + \Wv \mQ \mW_c)\rangle \end{bmatrix} + \lambda \nabla^2 \regularizer(\vtheta) \vd.
\end{equation}
Thus the total time complexity of \eqref{eq:final_Gd} is
\begin{equation*}
  \bbO{|\obs|k + (m+n)k^2+ 2m \costofforward{f} + 2n \costofforward{g}},
\end{equation*}
where the term `2' comes from the operations in \eqref{eq:Jvp} and \eqref{eq:ln_Gd0}. More details are in Section \ref{sec:imp}.

\section{Optimization Methods for Extreme Similarity Learning with Nonlinear Embeddings}
\label{sec:eom}
With the efficient computation of key components given in Section \ref{sec:fw}, we demonstrate that many optimization algorithms can be used. While block-wise minimization is applicable, we mentioned in Section \ref{sec:from_linear} that without a block-convex $\La$, such a setting may not be advantageous. Thus in our discussion we focus more on standard optimization techniques that update all variables at a time.

\subsection{(Stochastic) Gradient Descent Method}
\label{sec:GDM}
The standard gradient descent method considers the negative gradient direction 
\begin{equation}
  \vs = -\grad
\label{eq:grad_dir}
\end{equation}
and update $\vtheta$ by 
\begin{equation}
  \vtheta \gets \vtheta + \delta \vs,
 \label{eq:update_rule}
\end{equation}
where $\delta$ is the step size. To ensure the convergence, usually a line-search procedure is conducted to find $\delta$ satisfying the following sufficient decrease condition
\begin{equation}
  L(\vtheta + \delta \vs) \leq L(\vtheta)+\eta \delta \vs^\top \grad,
\label{eq:ls_condition}
\end{equation}
where $\eta \in (0,1)$ is a pre-defined constant. During line search, given any $\vtheta + \delta \vs$, the corresponding function value must be calculated.

Combined with the cost of $\grad$ in \eqref{eq:grad_comp} and the cost of $\La$ in \eqref{eq:com_fun_val}, the complexity of each gradient descent iteration is  
\begin{equation}
\begin{aligned}
  \bbO{(\LS + 1) &\times (|\obs|k + (m+n)k^2+ m \costofforward{f}+n \costofforward{g})},
\end{aligned}
\label{eq:gd_comp}
\end{equation}
where the term `1' in $(\LS + 1)$ comes from the cost of $\grad$, and \LS is the number of line search steps.

In some machine learning applications, the gradient calculation is considered too expensive so an approximation using partial data samples is preferred. This leads to the stochastic gradient (SG) method. Without the full gradient, the update rule \eqref{eq:update_rule} may no longer lead to the decrease of the function value. Thus a line search procedure is not useful and the step size $\delta$ is often decided by some pre-defined rules or tuned as a hyper-parameter. 

Unfortunately, it is known that for extreme similarity learning, a direct implementation of SG methods faces the difficulty of $\bbO{mn}$ cost (e.g., \cite{HFY16a-short}). Now all $(i, j)$ pairs are considered as our data set  and the cost of each SG step is proportional to the number of sampled pairs. Thus to go over all or most pairs many SG steps are needed  and the cost is $\bbO{mn}$. In contrast, by techniques in Section \ref{sec:grad} to calculate the gradient in $\bbO{m+n}$ cost,  all pairs have been considered. A natural remedy is to incorporate techniques in Section \ref{sec:grad} to the SG method. Instead of randomly selecting pairs in an SG step, \cite{WK19a-short} proposed to select pairs in a set $\batch = \hat{\sU} \times \hat{\sV}$ where  $\hat{\sU} \subseteq \sU$ and $\hat{\sV} \subseteq \sV$. If $\hat{m} = |\hat{\sU}|$ and $\hat{n} = |\hat{\sV}|$, then by slightly modifying the derivation in \eqref{eq:final_grad}, calculating the subsampled gradient involves $\bbO{\hat{m}+\hat{n}}$ instead of $\bbO{\hat{m}\hat{n}}$ cost. Analogous to \eqref{eq:grad_comp}, the subsampled gradient on $\batch$ can be calculated in 
$$\bbO{|\hat{\obs}|k + (\hat{m}+\hat{n})k^2+ \hat{m} \costofforward{f}+\hat{n} \costofforward{g}},$$ 
where $\hat{\obs}$ is the subset of observed pairs falling in $\hat{\sU} \times \hat{\sV}$. Then the complexity of each data pass  of handling $mn$ pairs via $\frac{mn}{\hat{m}\hat{n}}$ SG steps is
\begin{equation}
\bbO{|{\obs}|k + (\frac{mn}{\hat{n}}+\frac{mn}{\hat{m}})k^2+ \frac{mn}{\hat{n}} \costofforward{f}+\frac{mn}{\hat{m}}\costofforward{g}}.
\label{eq:ss_comp}
\end{equation}
Thus the $\bbO{mn}$ cost is deducted by a factor.

We consider another SG method that has been roughly discussed in \cite{WK19a-short}. The idea is to note that if $\obs_{i, :} = \{j : (i, j) \in \obs\}$ and $\obs_{:, j} = \{i : (i, j) \in \obs\}$, then
\begin{equation}\label{eq:trans}
\begin{aligned}
\dbsum \wi \wj(\cdots) &= \msum \sum_{j' \in \obs_{i, :}} \nsum \sum_{i' \in \obs_{:, j}}\frac{\wi}{|\obs_{i, :}|}\frac{\wj}{|\obs_{:, j}|} (\cdots)\\ &=\issum \jssum \frac{\wi}{|\obs_{i, :}|}\frac{\wj}{|\obs_{:, j}|}(\cdots).
\end{aligned}
\end{equation}
Then instead of selecting a subset from $mn$ pairs, the setting becomes to select two independent sets $\batchb, \batchc \subset \obs$ for constructing the subsampled gradient, where $\batchb$ and $\batchc$ respectively corresponding to the first and the second summations in \eqref{eq:trans}. Therefore, the task of going over $mn$ pairs is replaced by selecting $\batchb \subset \obs$, and $\batchc \subset \obs$ at a time to cover the whole $\obs \times \obs$ after many SG steps. We leave details of deriving the subsampled gradient in Appendix \ref{sec:SG2}, and the complexity for the task comparable to going over all $mn$ pairs by standard SG methods is
%Instead of selecting selecting pairs in $\batch = \hat{\sU} \times \hat{\sV}$, we discuss another SG method considering to select $\batch = \batchc \otimes \batchb$, where $\batchb$, $\batchc$ are two subsets independently subsampled from $\obs$, and
%\begin{equation*}
%  \batchc \otimes \batchb = \{(i, j), (i', j), (i, j'), (i',j')\mid \forall (i, j') \in \batchb, \forall (i',j)\in \batchc\}.
%\end{equation*}
%A preliminary version of this method is roughly discussed in \cite{WK19a-short}, but detailed derivations are not given. By showing their considered objective function is a special case of $\La$, we extend this method for solving \eqref{eq:prob}.
%Its idea also roots from \eqref{eq:mn_trick} but the setting is not as direct as the above one of selecting pairs in $\batch = \hat{\sU} \times \hat{\sV}$. 
%Thus we leave details in Appendix \ref{sec:SG2}, while listing the following complexity per data pass
\begin{equation}
  \bbO{(\frac{|\obs|^2}{|\batchb|}+\frac{|\obs|^2}{|\batchc|})(k + k^2+ \costofforward{f}+\costofforward{g})}.
  \label{eq:sogram_comp1}
\end{equation}
A comparison with \eqref{eq:ss_comp} shows how \eqref{eq:sogram_comp1} might alleviate the $\bbO{mn}$ cost. This relies on $|\obs| \ll mn$ and some appropriate choices of $|\batchb|$ and $|\batchc|$. We refer to this method as the SG method based on \eqref{eq:trans}.

The above setting is further extended in \cite{WK19a-short} to Stochastic Online Gramian (SOGram), which applies a variance reduction scheme to improve the subsampled gradient. Details are in Appendix \ref{sec:sogram}.
%Another SG method is Stochastic Online Gramian (SOGram) is proposed in \cite{WK19a-short}. By forming it as an extension of the SG method selecting $\batch = \batchc \otimes \batchb$, we extend it for solving \eqref{eq:prob}.
%which improves the subsampled gradient vector by considering a variance reduction scheme on estimates of \eqref{eq:gramian2} from a limited $\batch$. 
%We leave details of this method in Appendix \ref{sec:sogram}.
%where $\batchb$ and $\batchc$ are two subsets independently sampled from $\obs$; see details in Section \ref{sec:sogram}. 

We compare the above methods in Table \ref{tab:comp} by checking the cost per data pass, which roughly indicates the process of handling $mn$ pairs.
 Note that the total cost also depends on the number of data passes. In experiments we will see the trade-off between the cost per data pass and the number of total passes.

%In unconstrained minimization, gradient descent methods are known to have slow convergence. One can consider accelerated gradient descent method \cite{YEN83a} or other variants. Though we do not show details, these methods are applicable here because they need only extra vector operations similar to \eqref{eq:update_rule}. Note that each of these vector operations costs $\bbO{D}$, where $D = |\vtheta|$ is the number of parameters. This is in general cheaper than $\costofforward{f}+\costofforward{g}$  for generating nonlinear embeddings.

\subsection{Methods Beyond Gradient Descent}
\label{sec:sm}
We can further consider methods that incorporate the second-order information. By considering the following second-order Taylor expansion
\begin{equation}
  L(\vtheta + \vs) \approx L(\vtheta) + \grad^\top \vs + \half \vs^\top \hess \vs,
 \label{eq:general_taylor}
\end{equation}
Newton methods find a search direction $\vs$ by minimizing \eqref{eq:general_taylor}. However, this way of finding $\vs$ is often complicated and expensive. In the rest of this section, we discuss some practically viable options to use the second-order information.

\subsubsection{Gauss-Newton Method}
\label{sec:GN}
It is mentioned in Section \ref{sec:gnmvp} that Gauss-Newton matrix $\mG$ in \eqref{eq:general_G} is a positive-definite approximation of $\hess$. Thus replacing $\hess$ in \eqref{eq:general_taylor} with $\mG$ leads to a convex quadratic optimization problem with the solution satisfying  
\begin{equation}
  \mG \vs = -\grad.
\label{eq:general_ls}
\end{equation}
However, for large-scale applications, not only is $\mG$ in \eqref{eq:general_G} too large to be stored in memory, but the matrix inversion is also computationally expensive. 

Following \cite{CJL07b-short, JM10a-short}, we may solve \eqref{eq:general_ls} by the conjugate gradient (CG) method \citep{MRH52a-short}, which involves a sequence of matrix-vector products $\mG\vd$ between $\mG$ and some vector $\vd$. The derivation in Section \ref{sec:gnmvp} has shown how to calculate $\mG\vd$ without explicitly forming and storing $\mG$. Therefore, Gauss-Newton methods are a feasible approach for extreme similarity learning without the $\bbO{mn}$ cost.

After a Newton direction $\vs$ is obtained, the line search procedure discussed in Section \ref{sec:GDM} is conducted to find a suitable $\delta$ to ensure the convergence.
The complexity of each Newton iteration is 
\begin{equation}
\begin{aligned}
  \bbO{ \CG &\times (|\obs|k + (m+n)k^2+ 2m \costofforward{f} + 2n \costofforward{g}) \\
    +(\LS + 1) &\times (|\obs|k + (m+n)k^2+ m \costofforward{f}+n \costofforward{g})},
\end{aligned}
\label{eq:gn_comp}
\end{equation}
where \CG is numbers of CG steps. The overall procedures of the CG method and the Gauss-Newton method are summarized in Appendix \ref{sec:D_GN}. 

\subsubsection{Diagonal Scaling Methods}
\label{sec:DSM}
From \eqref{eq:gn_comp}, the CG procedure may be expensive so a strategy is to replace  $\mG$ in \eqref{eq:general_ls} with a positive-definite diagonal matrix, which becomes a very loose approximation of $\hess$. In this case, $\vs$ can be efficiently obtained by dividing each value in $-\grad$ by the corresponding diagonal element. Methods following this idea are referred to as diagonal scaling methods.

For example, the AdaGrad algorithm \cite{JD10a-short} is a popular diagonal scaling method. Given $\grad$ of current $\vtheta$, AdaGrad maintains a zero-initialized diagonal matrix $M\in \sR^{D \times D}$ by 
\begin{equation*}
  \mM \leftarrow \mM +  \diag(\grad)^2.
\end{equation*}
Then by considering $(\mu\mI+ \mM)^{\frac{1}{2}}$, where $\mI$ is the identity matrix and $\mu > 0$ is a small constant to ensure the invertibility, the direction is 
\begin{equation}
 \vs =  -(\mu\mI+ \mM)^{-\frac{1}{2}}\grad.
\label{eq:diag_scaling}
\end{equation}
In comparison with the gradient descent method, the extra cost is minor in $\bbO{D}$, where $D = D_u+D_v$ is the total number of variables. Instead of using the full $\grad$, this method can be extended to use subsampled gradient by the SG framework discussed in Section \ref{sec:GDM}.

\begin{table}
  \caption{A comparison of gradient-based methods on time complexity of each data pass. For the last setting, $|\batchh| = \min(|\batchb|, |\batchc|)$.}
  \centering
  \addtolength{\tabcolsep}{-4.5pt}
  \begin{tabular}{c|rrrrrrr}
  \toprule
    Method & \multicolumn{7}{c}{Time complexity per data pass} \\
  \midrule\midrule
    Naive SG &  & &  & & $mn \costofforward{f}$ & $+$ & $mn \costofforward{g}$ \\
  \midrule
%    GD, Newton & \multirow{2}{*}{$|\obs|k$} & \multirow{2}{*}{$+$} & \multirow{2}{*}{$(m+n)k^2$} & \multirow{2}{*}{$+$} & \multirow{2}{*}{$m \costofforward{f}$} & \multirow{2}{*}{$+$} & \multirow{2}{*}{$n \costofforward{g}$} \\
%    AGD, L-BFGS &  & & &  & & & \\
    Gradient (Alg. \ref{alg:grad-detailed}) & $|\obs|k$ & $+$ & $(m+n) k^2$ & $+$ & $m\costofforward{f}$ & $+$ & $n\costofforward{g}$ \\
    Selecting $\batch = \hat{\sU} \times \hat{\sV}$ & $|\obs|k$ & $+$ & $(\frac{n}{\hat{n}}m + \frac{m}{\hat{m}}n)k^2$ & $+$ & $\frac{n}{\hat{n}}m \costofforward{f}$ & $+$ & $\frac{m}{\hat{m}}n \costofforward{g}$ \\
%    Selecting $\batch = \batchb \otimes \batchc$ 
    SG based on \eqref{eq:trans} & $\frac{|\obs|^2}{|\batchh|}k$ & $+$ & $\frac{|\obs|^2}{|\batchh|} k^2$ & $+$ &
    $\frac{|\obs|^2}{|\batchh|}\costofforward{f}$ & $+$ & $\frac{|\obs|^2}{|\batchh|}\costofforward{g}$ \\
  \bottomrule
  \end{tabular}
  \label{tab:comp}
\end{table}

\section{Implementation Details}
\label{sec:imp}

So far we used $\costofforward{f}$ and $\costofforward{g}$ to represent the cost of operations on nonlinear embeddings.
Here we discuss more details if automatic differentiation is employed for easily trying different network architectures.
To begin, note that $[\xJfi, \xJgj]$ is the Jacobian of $(\Fi, \Fj)$.
Transposed Jacobian-vector products are used in $\grad$ for all methods and $\mG \vd$ for the Gauss-Newton method.
In fact, \eqref{eq:final_grad} and \eqref{eq:final_Gd} are almost in the same form.
It is known that for such an operation, the reverse-mode automatic differentiation (i.e., back-propagation) should be used so that only one pass of the network is needed \citep{AGB18a-short}.
For feedforward networks, including extensions such as convolutional networks, the cost of a backward process is within a constant factor of the forward process for computing the objective value (e.g., Section 5.2 of \cite{CCW18a-short}).
The remaining major operation to be discussed is the Jacobian-vector product in \eqref{eq:Jvp} in Section \ref{sec:GN}.
This operation can be implemented with the forward-mode automatic differentiation in just one pass \citep{AGB18a-short}, and the cost is also within a constant factor of the cost of forward objective evaluation.\footnote{See, for example, the discussion at \url{https://www.csie.ntu.edu.tw/~cjlin/courses/optdl2021/slides/newton_gauss_newton.pdf}.}
We conclude that if neural networks are used, using $\costofforward{f}$ and $\costofforward{g}$ as the cost of nonlinear embeddings is suitable.

Neural networks are routinely trained on GPU instead of CPU for better efficiency, but the smaller memory of GPU causes difficulties for larger problems or network architectures.
This situation is even more serious for our case because some other matrices besides the two networks are involved; see \eqref{eq:final_grad} and \eqref{eq:final_Gd}.
We develop some techniques to make large-scale training feasible, where details are left in supplementary materials.

\section{Experiments}
\label{sec:exp}
In this section, after presenting our experimental settings, we compare several methods discussed in Section \ref{sec:eom} in terms of the convergence speed on objective value decreasing and achieving the final test performance.

\subsection{Experimental Settings}
\subsubsection{Data sets} We consider three data sets of recommender systems with implicit feedback and one data set of link predictions. 
For data sets of recommender systems with implicit feedback, the target similarity of each observed user-item pair is 1, while those of the unobserved pairs are unrevealed. We consider \mlone, \mlten, and \net provided by \cite{HFY16a-short}. 
For the data set of link predictions, we follow \cite{WK19a-short} to consider the problem of learning the intra-site links between Wikipedia pages, where the target similarity of each observed $\text{page}_\text{from}$-$\text{page}_\text{to}$ pair is 1 if there is a link from $\text{page}_\text{from}$ to $\text{page}_\text{to}$, and unrevealed otherwise. 
However, because the date information of the dump used in \cite{WK19a-short} is not provided, their generated sets are irreproducible.
We use the latest Wikipedia graph\footnote{\url{https://dumps.wikimedia.org/simplewiki/20200901/}} on pages in simple English and denote the generated data set as \wiki. For \mlone, \mlten and \net, training and test sets are available. For \wiki, we follow \cite{WK19a-short} to use a 9-to-1 ratio for the training/test split.
The statistics of all data sets are listed in \Tabref{tab:data}. 
Note that the aim here is to check the convergence behavior, so we do not need a validation set for hyper-parameter selection.

\subsubsection{Model and Hyper-parameters}
\label{sec:mhp}
We train a two-tower neural network as in \Figref{fig:twin}. Both towers have three fully-connected layers, where the first two layers contain 256 hidden units equipped with ELU \citep{CD16a-short} activation, and the last layer contains $k=128$ hidden units without activation.
For all experiments,
we choose the logistic loss $\ell(y, \hat{y}) = \log(1 + \exp(-y \hat{y}))$,
an L2 regularizer $\regularizer(\vtheta) = \frac{1}{2}\|\vtheta\|_2^2$, and
uniform weights $\wi = 1, \forall i = 1, \dots, m, \text{ } \wj = 1, \forall j = 1, \dots, n$ in \eqref{eq:loss}.
%embedding dimension $k = 128$.
For the imputed labels $\tilde{\mY}$, we follow past works \citep{HFY16a-short, HFY17a-short} to apply a constant $-1$ for any unobserved pair by  
setting $\tZi = -\boldsymbol{1}/\sqrt{k}$, and $\tZj = \boldsymbol{1}/\sqrt{k}$.
For $\nw$ and $\lambda$, for each data set, we consider a grid of hyper-parameter combinations and select the one achieving the best results on the test set. 
The selected $\nw$ and $\lambda$ for each data set are listed in \Tabref{tab:data}.

\begin{figure*}
%  \begin{subfigure}[b]{\textwidth}
%    \centering
%    \includegraphics[width=0.25\linewidth]{./figs/ml1m-step-f.eps}
%    \hspace{-0.3cm}
%    \includegraphics[width=0.25\linewidth]{./figs/ml10m-step-f.eps}
%    \hspace{-0.3cm}
%    \includegraphics[width=0.25\linewidth]{./figs/netflix-step-f.eps}
%    \hspace{-0.3cm}
%    \includegraphics[width=0.25\linewidth]{./figs/wiki-simple-step-f.eps}
%    \caption{Relative difference in objective value versus the number of steps.} \label{fig:step-f}
%  \end{subfigure}
%  \vskip\baselineskip
  \begin{subfigure}[b]{\textwidth}
    \centering
    \includegraphics[width=0.25\linewidth]{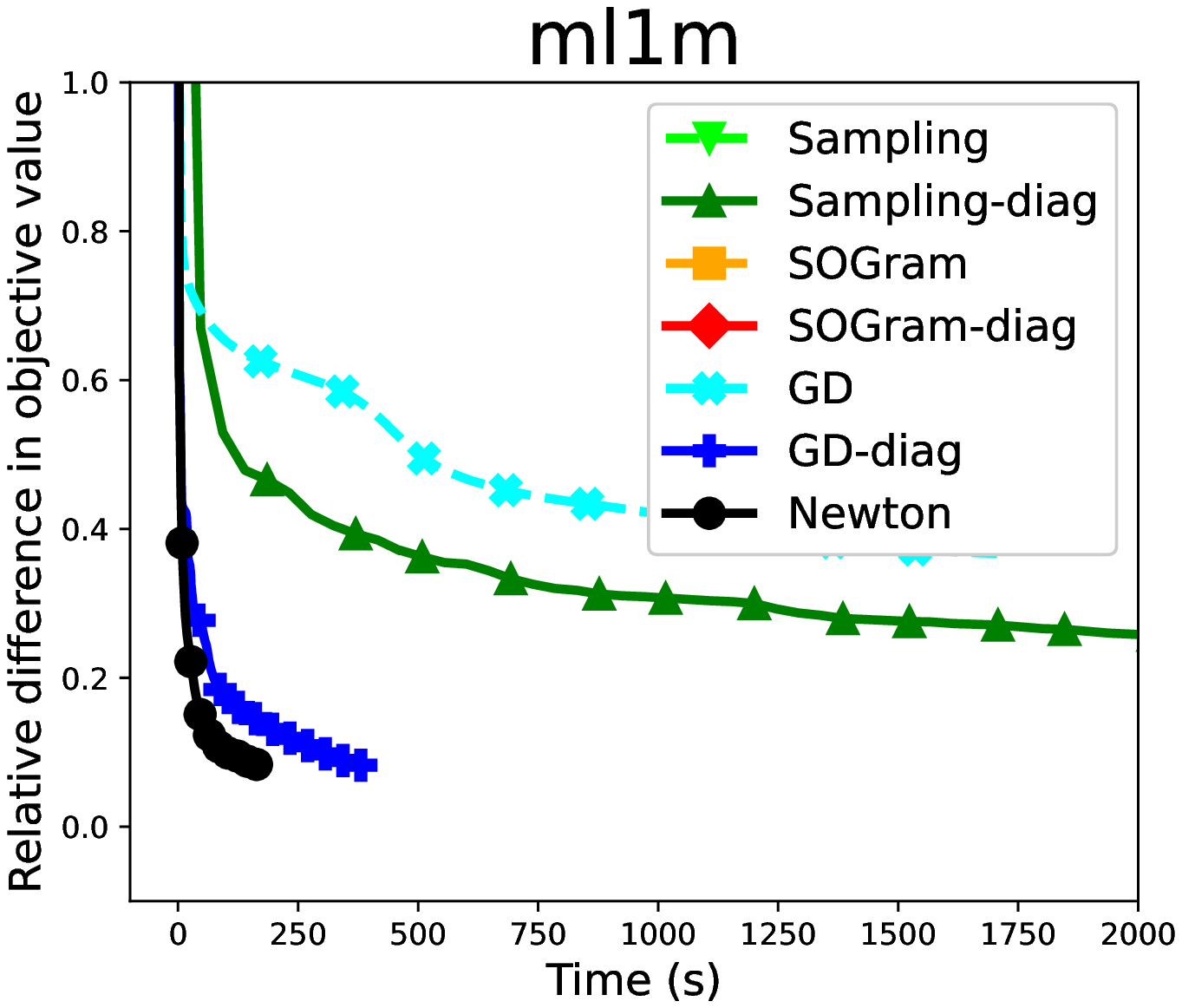}
    \hspace{-0.3cm}
    \includegraphics[width=0.25\linewidth]{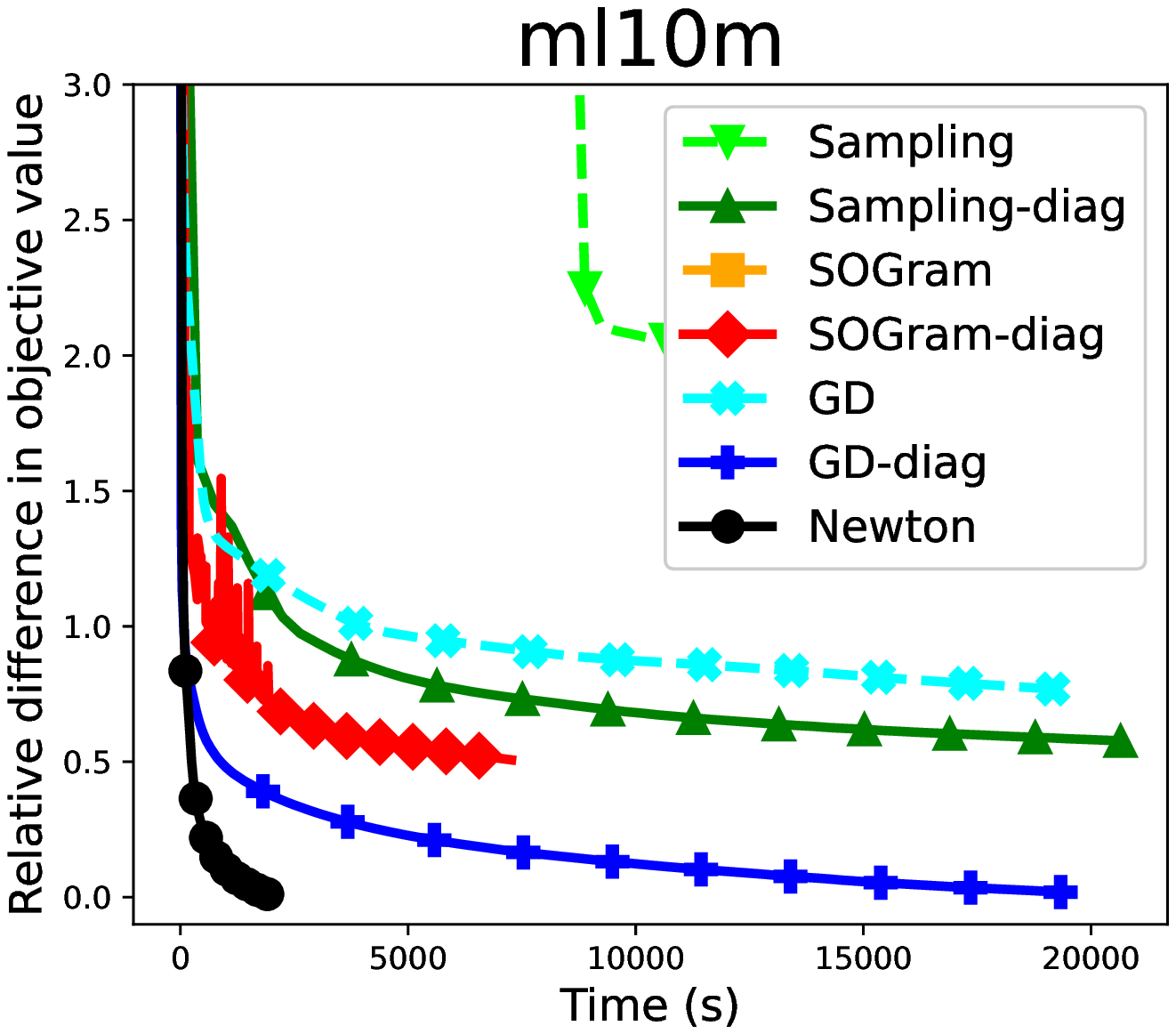}
    \hspace{-0.3cm}
    \includegraphics[width=0.25\linewidth]{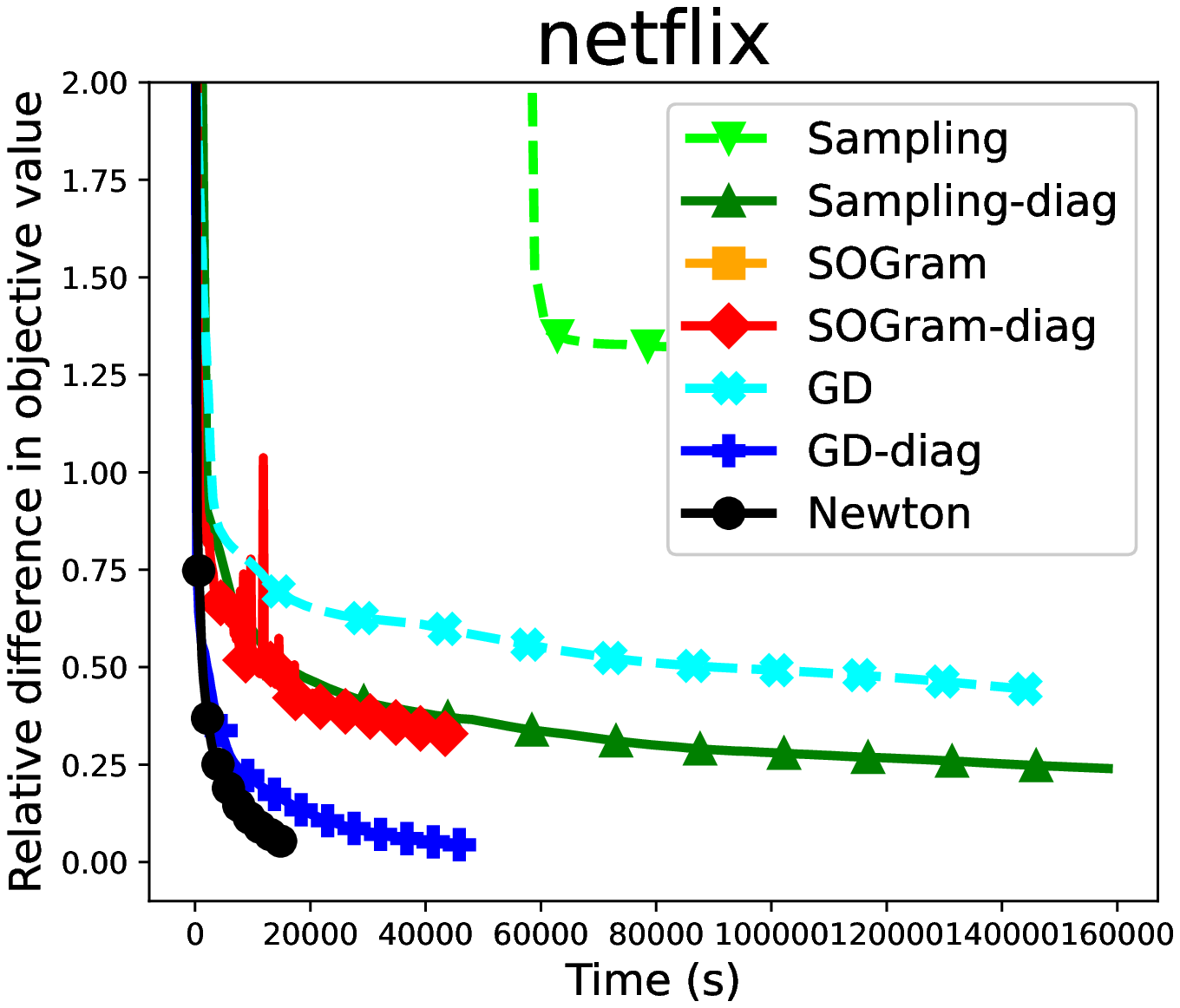}
    \hspace{-0.3cm}
    \includegraphics[width=0.25\linewidth]{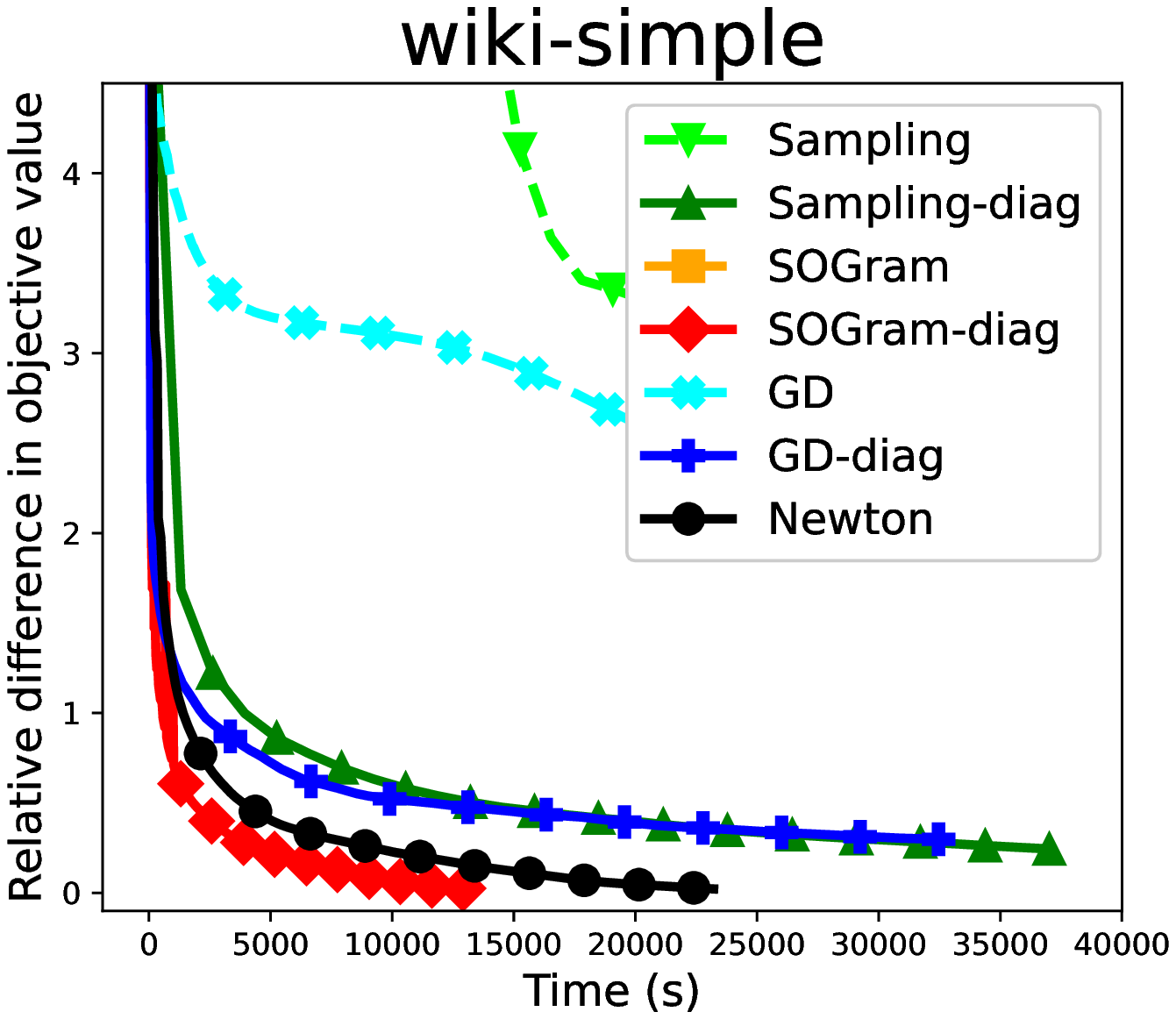}
    \caption{Relative difference in objective value versus training time.} \label{fig:time-f}
  \end{subfigure}
  \vskip\baselineskip
  \begin{subfigure}[b]{\textwidth}
    \centering
    \includegraphics[width=0.25\linewidth]{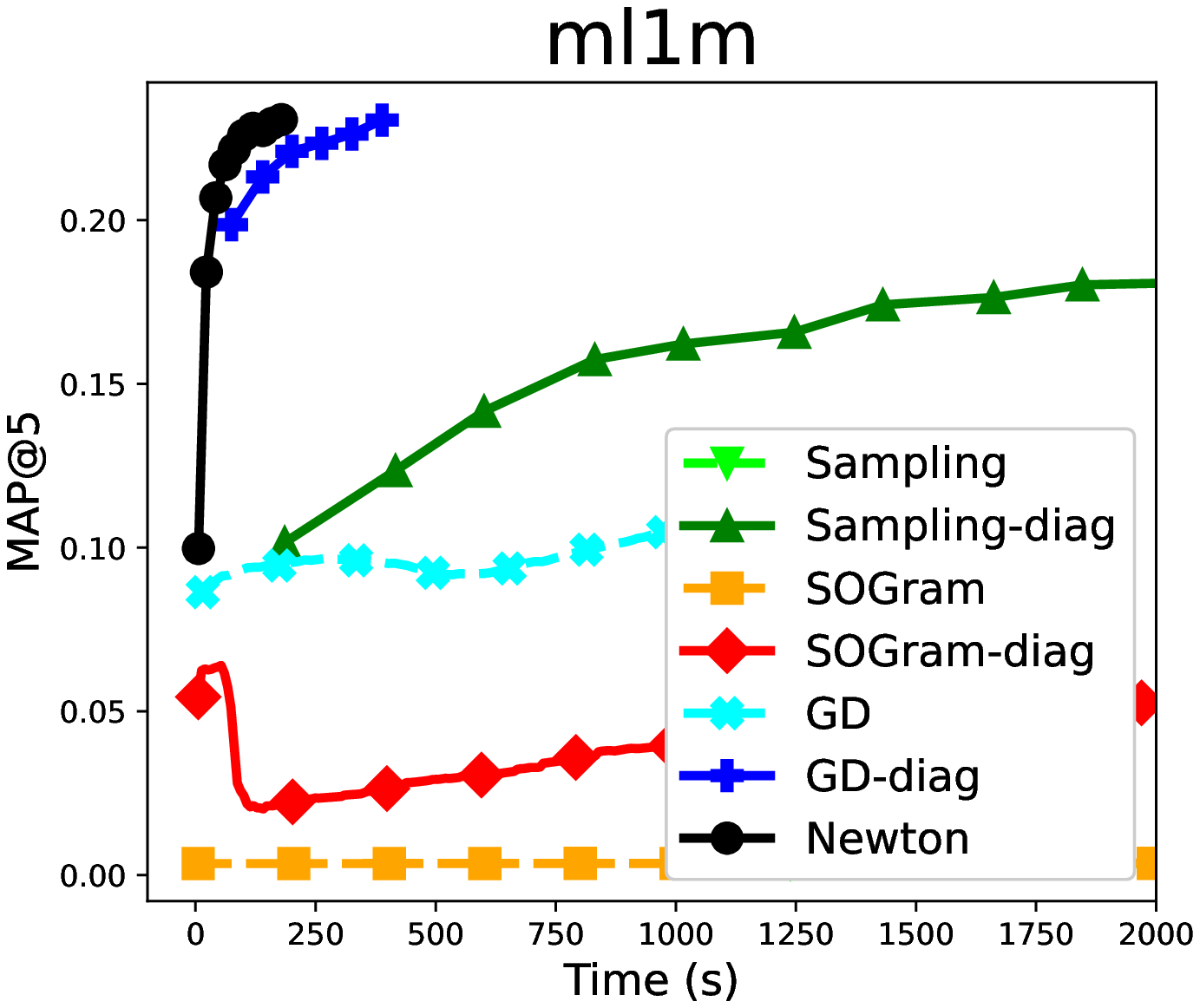}
    \hspace{-0.3cm}
    \includegraphics[width=0.25\linewidth]{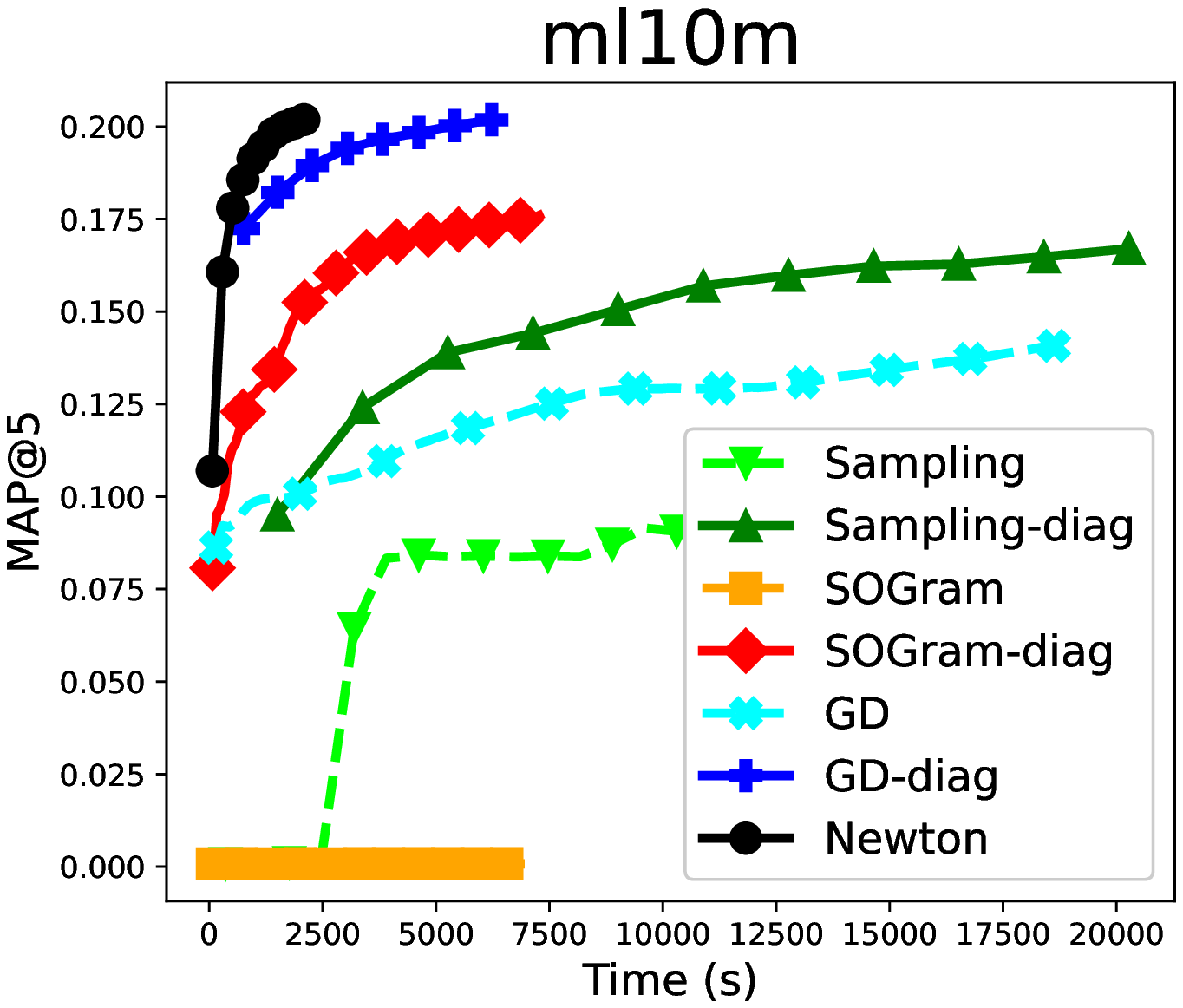}
    \hspace{-0.3cm}
    \includegraphics[width=0.25\linewidth]{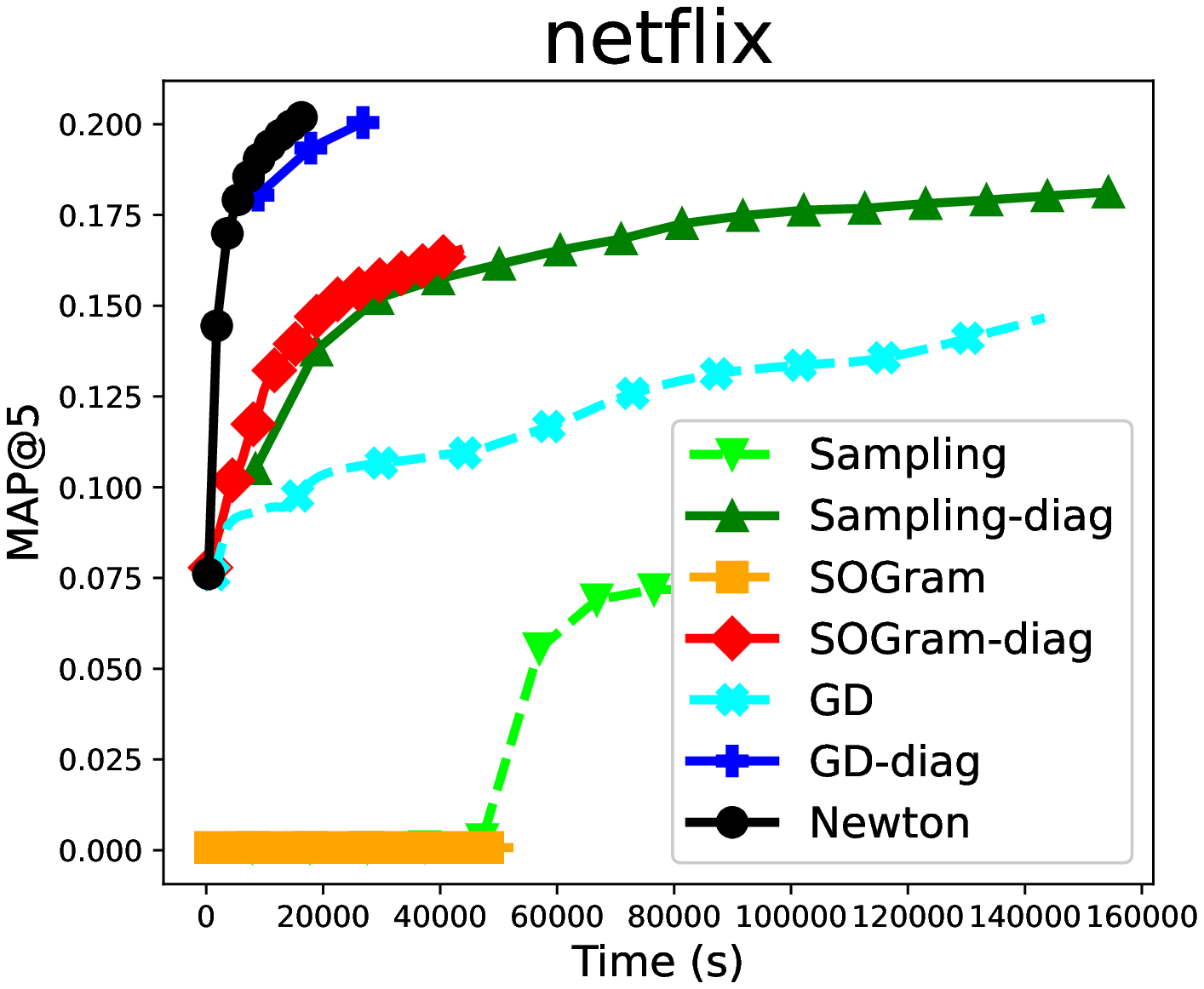}
    \hspace{-0.3cm}
    \includegraphics[width=0.25\linewidth]{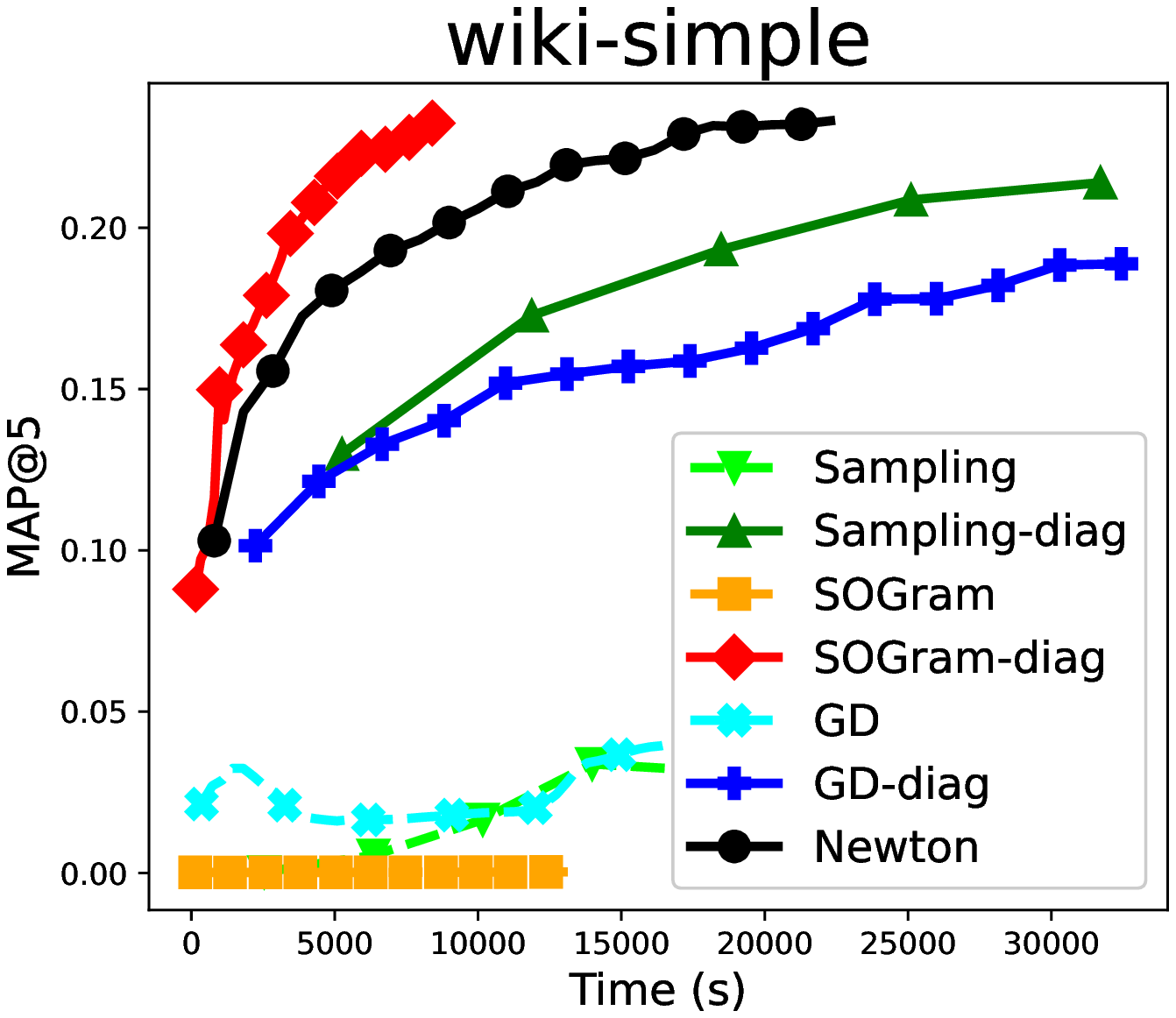}
    \caption{MAP@5 versus training time.} \label{fig:time-map}
  \end{subfigure}
  \caption{Comparison of different algorithms on four data sets. \mbsogram and \mbss may be too slow to be shown.}
  \Description{Comparison of different algorithms on four data sets.}
  \label{fig:perf}
\end{figure*}

\begin{table}
\caption{Data statistics and the selected hyper-parameters.}
\centering
% \vskip 1em
\begin{tabular}{l|rrrrr}
\toprule
  Data set & $m$ & $n$ & $|\obs|$ &  $\log_2\nw$ & $\log_2\lambda$ \\
\midrule
  \mlone & 6,037 & 3,513 & 517,770 & $-4$ & $2$ \\
  \mlten & 69,786 & 10,210 & 4,505,820  & $-8$ & $-2$ \\
  \net & 478,251 & 17,768 & 51,228,351 & $-8$ & $0$ \\
  \wiki & 85,270 & 55,695 & 5,478,549 & $-10$ & $2$ \\
\bottomrule
\end{tabular}
\label{tab:data}
\end{table}

\subsubsection{Compared Optimization Methods}
\label{sec:COM}
We compare seven optimization methods for extreme similarity learning, which are categorized into two groups. 
The first group includes the following three gradient-based methods discussed in Section \ref{sec:GDM}.
\begin{itemize}[leftmargin=*]
 \item \gd: This is the gradient descent method. For each step, we compute $\vs = -\grad$ by \eqref{eq:final_grad}.  To ensure the convergence, backtracking line search is conducted to find the step size $\delta$.
  \item \mbsogram: This is a stochastic gradient method proposed in \cite{WK19a-short}; see the discussion in Appendix \ref{sec:sogram}. 
  We set $|\batchb| = |\batchc| = \rho|\obs|$, where $\rho = 10^{-2}$
  is the sampling ratio of observed pairs.
  We follow \cite{WK19a-short} to set $\alpha = 0.1$, which is a hyper-parameter of controlling the bias-variance tradeoff.
  \item \mbss: This is the stochastic gradient method of choosing $\batch = \hat{\sU} \times \hat{\sV}$ \cite{WK19a-short}.
  To make the number of steps in each data pass comparable to \mbsogram, for \mbss we set $\hat{m} = \rho m$ and $\hat{n} = \rho n$, where $\rho$ is the ratio used in \mbsogram.
  \end{itemize}
The step size $\delta$ in \mbsogram and \mbss is set to $2^{-25}$, $2^{-23}$, $2^{-25}$ and $2^{-25}$ respectively for \mlone, \mlten, \net and \wiki. These values are from \gd's most used $\delta$.

The second group includes the following four methods incorporating the second-order information discussed in Section \ref{sec:sm}.
  \begin{itemize}[leftmargin=*]
  \item \gn:
  This is the Gauss-Newton method proposed in Section \ref{sec:GN}.
  For CG, we set the stopping tolerance $\xi = 0.1$ in Appendix \Algref{alg:cg-detailed} and 30 as the maximal number of CG steps.
  \item \fullss, \fullsogram, \fullsg: These three methods are respectively extended from \gd, \mbsogram, and \mbss by applying the diagonal scaling in \eqref{eq:diag_scaling}. For \fullss, the same as \gd, it finds $\delta$ by conducting backtracking line search. For \fullsogram and \fullsg, we set $\delta=0.01$.  
 \end{itemize}
 For \gd, \gn, and \fullss, which conduct line search, we set $\eta = 10^{-4}$ for the sufficient decrease condition \eqref{eq:ls_condition} and sequentially check $\delta$ in $\{\delta_\text{init}, \frac{\delta_\text{init}}{2}, \frac{\delta_\text{init}}{4},\dots\}$, where $\delta_\text{init}$ is the step size taken in the last iteration, and is doubled every five iterations to avoid being stuck with a small step size.
% For the stopping condition, we terminate them when they achieve the best test performance or the running time reaches ten times of that of the fastest method.

\subsubsection{Environment and Implementation}
We use TensorFlow \citep{MA16a-short} compiled with Intel\textregistered{ } Math Kernel Library (MKL) to implement all algorithms.
Because most operations are matrix-based and MKL is optimized for such operations, our implementation should be sufficiently efficient. For sparse operations involving iterating $\obs$ (e.g., $\Lp$ and $\mX \mQ$), we implement them by NumPy \citep{CRH20a} and a C extension, where we parallelize these operations by OpenMP. 
As our applied neural networks described in Section \ref{sec:mhp} are not complex, we empirically do not observe many advantages of GPU over CPU on the running time.
Thus all experiments are conducted on a Linux machine with 10 cores of Intel\textregistered{ }  Core\texttrademark{ } i7-6950X CPUs and 128GB memory.

\subsubsection{Evaluation Criterion of Test Performance}
On test sets, we report mean average precision (MAP) on top-$5$ ranked entities.
For left entity $i$, let $\tilde{\obs}_{i, :}$ be the set of its similar right entities in the test set, and $\tilde{\obs}^\text{sorted}_{i, :K}$ be the set of top-$K$ right entities with the highest predicted similarity.  We define
\begin{equation}
  \mathrm{MAP}@5 = \frac{1}{5} \sum_{K=1}^5 \frac{1}{\tilde{m}} \sum_{i=1}^{\tilde{m}} \frac{1}{K} |\tilde{\obs}_{i, :} \cap \tilde{\obs}^\text{sorted}_{i, :K}|,
\label{eq:map}
\end{equation}
where $\tilde{m}$ is the number of left entities in the test set. 

\subsection{A Comparison on the Convergence Speed} \label{sec:compare}

We first investigate the relationship between the running time and the relative difference in objective value $$(L(\vtheta) - L^*) / L^*,$$ where $L^*$ is the lowest objective value reached among all settings. The result is presented in \Figref{fig:time-f}.

For the three gradient-based methods, \mbsogram, and \mbss are slower than \gd. This situation is different from that of training general neural networks, where SG are the dominant methods.
The reason is that though \mbss and \mbsogram take more iterations than \gd in each data pass, from \Tabref{tab:comp}, they have much higher complexity per data pass. 
%By contrast, for training general neural networks, SG methods generally have the same complexity as the gradient descent method.

For \fullss, \fullsogram, \fullsg, and \gn, which incorporate the second-order information, they are consistently faster than gradient-based methods. Specifically, between \gn and \gd, the key difference is that \gn additionally considers the second-order information for yielding each direction.  The great superiority of \gn confirms the importance of leveraging the second-order information. 
By comparing \fullss and \gd, we observe that \fullss is much more efficient. Though both have similar complexities for obtaining search directions, \fullss leverages partial second-order information from AdaGrad's diagonal scaling. However, the effect of the diagonal scaling is limited, so \fullss is slower than \gn. Finally, \fullss is still faster than \fullsogram, \fullsg as in the case of without diagonal scaling.

Next, we present in \Figref{fig:time-map} the relationship between the running time and MAP@5 evaluated on test sets. It is observed that a method with faster convergence in objective values is also faster in MAP@5.

\section{Conclusions}
In this work, we study extreme similarity learning with nonlinear embeddings. The goal is to alleviate the $\bbO{mn}$ cost in the training process. While this topic has been well studied for the situation of using linear embeddings, a systematic study for nonlinear embeddings was lacking. We fill the gap in the following aspects. First, for important operations in optimization algorithms such as function and gradient evaluation, clean formulations with $\bbO{m+n}$ cost are derived. Second, these formulations enable the use of many optimization algorithms for extreme similarity learning with nonlinear embedding models. We detailedly discuss some of them and check implementation issues. Experiments show that efficient training by some algorithms can be achieved.
\label{sec:con}

\begin{acks}
We would like to thank Hong Zhu, Yaxu Liu, and Jui-Nan Yen for helpful discussions. This work was supported by MOST of Taiwan grant 107-2221-E-002-167-MY3.
\end{acks}

\bibliographystyle{ACM-Reference-Format}
\bibliography{sdp}

\clearpage
\appendix
\DeclareRobustCommand{\vect}[1]{\bm{#1}}
\pdfstringdefDisableCommands{%
  \renewcommand{\vect}[1]{#1}%
}

\section{Appendix}
\subsection{Detailed Derivation of $L^-(\vect{\theta})$}
\label{sec:D_fun_val}
To compute $\Ln$, we rewrite it as
\begin{align}
  &\Ln=\half \dbsum \wi\wj (\tZi^\top\tZj-\Zi^\top \Zj)^2 \nonumber \\
  =&\half \dbsum \wi\wj \Big(\tZi^\top \tZj (\tZj^\top \tZi)-2\tZi^\top\tZj(\Zj^\top\Zi)+\Zi^\top \Zj (\Zj^\top \Zi)\Big) \nonumber\\
  =&\half \msum \Big(\wi \tZi^\top(\nsum \wj \tZj\tZj^\top)\tZi - 2 \wi \tZi^\top(\nsum \wj\tZj\Zj^\top)\Zi \nonumber \\
  &+ \wi \Zi^\top(\nsum \wj \Zj\Zj^\top)\Zi\Big) \nonumber \\
   =&\half \fip{\tPc}{\tQc}-\fip{\hPc}{\hQc}+ \half \fip{\Pc}{\Qc}.\label{eq:final_fun}\end{align}
To calculate \eqref{eq:final_fun}, we need matrices listed in \eqref{eq:gramian}. For $\tPc$ and $\tQc$, they are constant matrices throughout the training process, so we can calculate them and $\fip{\tPc}{\tQc}$ within $\bbO{(m+n)k^2}$ at the beginning. Thus this cost can be omitted in the complexity analysis.
For other matrices in \eqref{eq:gramian}, $\mP$ and $\mQ$ are needed,
but they have been pre-computed and cached in memory during computing $\Lp$.
Thus all matrices in \eqref{eq:gramian} can be obtained in $\bbO{(m+n)k^2}$ time and cached in $\bbO{k^2}$ space. Then in \eqref{eq:final_fun}, the Frobenius inner products between these matrices cost only $\bbO{k^2}$ time.
%The overall time complexity of evaluating $\La$ is 
%$$\bbO{|\obs|k + (m+n)k^2+ m \costofforward{f}+n \costofforward{g}}.$$
Through combining $\Lp$ calculation discussed in Section \ref{sec:funval}, \Algref{alg:funval-detailed} summarizes the procedure of evaluating the objective function.

\subsection{Detailed Derivation of a Stochastic Gradient Method Based on \eqref{eq:trans}}
\label{sec:SG2}
From \eqref{eq:trans},
by defining $\gwi = \frac{\wi}{|\obs_{i, :}|}$ and $\gwj = \frac{\wj}{|\obs_{:, j}|}$, 
$\La$ in \eqref{eq:obj} can be written as
\begin{equation}\label{eq:sogram_obj}
\begin{aligned}
& \reg\regularizer(\vtheta) + \ssum \ell^+_{ij}(\Yij, \hYij)
+ \omega \issum \jssum \frac{1}{2}\gwi \gwj (\tYij-\hYij)^2.
\end{aligned}
\end{equation}
Then $\ngrad$ in \eqref{eq:ln_grad_init} can be rewritten as
\begin{align}
\ngrad=&\issum\jssum \begin{bmatrix} \tJfi\Zj \\ \tJgj\Zi \end{bmatrix} \gwi\gwj (\Zi^\top\Zj-\tZi^\top \tZj) \label{eq:sogram_ngrad_init}\\
=& \begin{bmatrix}
  \issum \tJfi(\Qo\gwi\Zi - \hQo^\top\gwi\tZi) \\
  \jssum \tJgj(\Po\gwj\Zj - \hPo ^\top\gwj\tZj)
\end{bmatrix} \label{eq:sogram_ngrad},
\end{align}
where similar to \eqref{eq:gramian}, we define
\begin{equation}
\begin{aligned}
\Po &= \issum \gwi\Zi\Zi^\top, \quad
\Qo = \jssum \gwj\Zj\Zj^\top, \\
\hPo &= \issum \gwi\tZi\Zi^\top, \quad
\hQo = \jssum \gwj\tZj\Zj^\top.
\end{aligned}
\label{eq:gramian_full}
\end{equation}
and we omit details from \eqref{eq:sogram_ngrad_init} to \eqref{eq:sogram_ngrad} as the derivations are similar to those between \eqref{eq:ln_grad_init} and  \eqref{eq:ln_grad}.

On the other hand, from \eqref{eq:Xij} and \eqref{eq:tJyij}, we write $\pgrad$ as
\begin{equation}
  \pgrad =  \begin{bmatrix}
    \ssum \Jfi^\top \Zj X_{ij} \\
    \ssum \Jgj^\top \Zi X_{ij}
  \end{bmatrix}.
\label{eq:sogram_pgrad}
\end{equation}
Because both \eqref{eq:sogram_ngrad} and  \eqref{eq:sogram_pgrad} are now operations summing over $\obs$, $\grad$ can be written as
\begin{align}
\begin{bmatrix}
  \ssum \Jfi^\top \big(\Zj X_{ij} + \nw (\Qo\gwi\Zi - \hQo^\top\gwi\tZi) \big) \\
  \ssum \Jgj^\top \big(\Zi X_{ij} + \nw (\Po\gwj\Zj - \hPo^\top\gwj\tZj) \big)
\end{bmatrix}+\lambda \nabla \regularizer(\vtheta).
\label{eq:grad_sogram_full}
\end{align}
%Let $\batchb$, and $\batchc$ be two subsets independently subsampled from $\obs$.
To derive a subsampled approach, we must consider two subsets $\batchb \subset \obs$ and $\batchc \subset \obs$ because from \eqref{eq:gramian_full}, two summations respectively over $\obs$ are involved in the above gradient. This results in the following estimate of $\grad$.
\begin{equation}
\frac{|\obs|}{|\batchb|} \begin{bmatrix}
  \bsumb \Jfi^\top \big(\Zj X_{ij} + \nw (\Qbc\gwi\Zi - \hQbc^\top\gwi\tZi) \big) \\
  \bsumb \Jgj^\top \big(\Zi X_{ij} + \nw (\Pbc\gwj\Zj - \hPbc^\top\gwj\tZj) \big)
\end{bmatrix}+\lambda \nabla \regularizer(\vtheta),
\label{eq:grad_sogram_naive}
\end{equation}
where 
\begin{equation}\label{eq:gramian2}
\begin{aligned}
\Pbc &= \frac{|\obs|}{|\batchc|} \bsumc \gwi\Zi\Zi^\top, \quad
\Qbc = \frac{|\obs|}{|\batchc|} \bsumc \gwj\Zj\Zj^\top, \\
\hPbc &= \frac{|\obs|}{|\batchc|} \bsumc \gwi\tZi\Zi^\top, \quad
\hQbc = \frac{|\obs|}{|\batchc|} \bsumc \gwj\tZj\Zj^\top.
\end{aligned}
\end{equation}
To make \eqref{eq:grad_sogram_naive} an unbiased estimate of $\grad$, we need that $\batchb$ and $\batchc$ are independent of each other, and have the scaling factors $\frac{|\obs|}{|\batchb|}$, $\frac{|\obs|}{|\batchc|}$ respectively in \eqref{eq:grad_sogram_naive} and \eqref{eq:gramian2}.

Because $\batchb$ and $\batchc$ are independent subsets, we can swap them in \eqref{eq:grad_sogram_naive} to have another estimate of $\grad$. Then four matrices $\Pbb,\Qbb,\hPbb$ and $\hQbb$ similar to those in \eqref{eq:gramian2} must be calculated, but the summations are now over $\batchb$. In the end, the two subsampled gradients can be averaged.

Similar to the discussion on \eqref{eq:gramian} and \eqref{eq:final_grad}, for each pair of $\batchb$ and $\batchc$, \eqref{eq:grad_sogram_full} can be computed in
\begin{equation} \label{eq:grad_comp_batch}
  \gO((|\batchb|+|\batchc|)(k + k^2 + \costofforward{f} + \costofforward{g}))
\end{equation} 
time.
We take $\bbO{|\batchc|k^2}$ term as an example to illustrate the difference from \eqref{eq:final_grad}. Now for matrices in \eqref{eq:gramian2}, each involves $\bbO{|\batchc|k^2}$ cost, but for those in \eqref{eq:gramian}, $\bbO{mk^2}$ or $\bbO{nk^2}$ are needed.
Therefore, from \eqref{eq:grad_comp} to \eqref{eq:grad_comp_batch}, we can see the $m+n$ term should be replaced by $|\batchb|+|\batchc|$.

Next we check the complexity.
We have mentioned in Section \ref{sec:GDM} that the task of going over all $mn$ pairs is now replaced by sampling $\batchb \times \batchc$ to cover $\obs \times \obs$.
Thus the time complexity is by multiplying the $\bbO{\frac{|\obs|^2}{|\batchb||\batchc|}}$ steps and the cost in \eqref{eq:grad_comp_batch} for each SG step to have
\begin{equation}
  \bbO{(\frac{|\obs|^2}{|\batchb|}+\frac{|\obs|^2}{|\batchc|})(k + k^2+ \costofforward{f}+\costofforward{g})}.
  \label{eq:sogram_comp}
\end{equation}

\begin{algorithm}[t]
    \caption{Objective function evaluation: $\La$}
    \label{alg:funval-detailed}
  \DontPrintSemicolon
	\KwIn{$\vtheta, \sU, \sV, \obs, \tilde{\mP}, \tilde{\mQ}, \Wu, \Wv, \fip{\tPc}{\tQc}$.}
%	$\Zi \gets \Fi, \Zj \gets \Fj$ \\
	$\mP \gets [\cdots f(\vtheta^u; \vu_i) \cdots]^\top, \mQ \gets [\cdots g(\vtheta^v; \vv_j) \cdots]^\top$\\
    $\Pc \gets \mP^\top \Wu \mP, \Qc \gets \mQ^\top \Wv \mQ$\\
    $\hPc \gets \tilde{\mP}^\top \Wu \mP, \hQc \gets \tilde{\mQ}^\top \Wv \mQ$ \\
    $\Lp \gets \ssum \lij^+(Y_{ij}, \Zi^\top \Zj)$\\
    $\Ln \gets \half \fip{\tPc}{\tQc}-\fip{\hPc}{\hQc}+ \half \fip{\Pc}{\Qc}$\\
    $\La \gets \Lp + \nw \Ln + \lambda \regularizer(\vtheta)$\\
    \KwOut{$\La$, $\mP, \mQ, \Pc, \Qc, \hPc, \hQc$} 
\end{algorithm}

\subsection{Detailed Derivation of the SOGram Method}
\label{sec:sogram}

We discuss the Stochastic Online Gramian (SOGram) method proposed in \cite{WK19a-short}, which is very related to the method discussed in Section \ref{sec:SG2}. It considers a special case of the objective function in \eqref{eq:sogram_obj} by excluding $\gwi$ and $\gwj$. From this we show that SOGram is an extension.

In \cite{WK19a-short}, they view matrices in \eqref{eq:gramian2} as estimates of matrices in \eqref{eq:gramian}.  The main idea behind SOGram is to
apply a variance reduction scheme to improve these estimates.
Specifically, they maintain four zero-initialized matrices with the following exponentially averaged updates before each step
\begin{equation}
\begin{aligned}
  \Pc'  &\gets (1-\alpha) \Pc'  + \alpha \Pbc , \quad
   \Qc'   \gets (1-\alpha) \Qc'  + \alpha  \Qbc , \\
  \hPc' &\gets (1-\alpha) \hPc' + \alpha  \hPbc, \quad
  \hQc'  \gets (1-\alpha) \hQc' + \alpha   \hQbc,
\end{aligned}
\label{eq:gramian_sogram}
\end{equation}
where $\alpha$ is a hyper-parameter for controlling the bias-variance tradeoff. 
Then they replace \eqref{eq:grad_sogram_naive} with
\begin{equation}
\frac{|\obs|}{|\batchb|}
\begin{bmatrix}
  \bsumb \Jfi^\top \Big(\Zj X_{ij} + \nw (\Qc'\gwi\Zi - \hQc'^\top\gwi\tZi) \Big) \\
  \bsumb \Jgj^\top \Big(\Zi X_{ij} + \nw (\Pc'\gwj\Zj - \hPc'^\top\gwj\tZj) \Big)
\end{bmatrix}+\lambda \nabla \regularizer(\vtheta),
\label{eq:grad_sogram}
\end{equation}
which is considered as the subsampled gradient vector for each SG step.

\begin{algorithm}[t]
    \caption{Conjugate gradient (CG) procedures}
    \label{alg:cg-detailed}
  \DontPrintSemicolon
	\KwIn{$\vtheta, \obs, \grad, \Wu, \Wv, \mP, \mQ, \Pc, \Qc$.}
      $\vr \gets -\grad$,
      $\vd \gets \vr$,
      $\vs \gets$\textbf{0},
      $\gamma \gets \vr^\top \vr$,
      $\gamma_0 \gets \gamma$\\
    \Repeat{$ \sqrt{\gamma} \leq \xi \sqrt{\gamma_0}$ \text{or the max number of CG steps is reached }}{
	  $\mW \gets [\cdots \Jfi\vd^u \cdots]^\top, \mH \gets [\cdots \Jgj\vd^v\cdots]^\top$\\
      $\mW_c \gets \mW^\top \Wu \mP, \mH_c \gets \mH^\top \Wv \mQ$ \\
	  $Z_{ij} \gets \sparti{\lij^+}{\hYij}(\bwi^\top\Zj + \Zi^\top\bhj), \forall (i, j)\in \obs$ \\
	  $\mZ_q \gets \mZ\mQ$, $\mZ_p \gets \mZ^\top \mP$\\
    $\mG\vd \gets \begin{bmatrix} \langle \tJf, \mZ_q + \nw(\Wu \mW \Qc + \Wu \mP \mH_c) \rangle \\ \langle \tJg, \mZ_p + \nw(\Wv \mH \Pc + \Wv \mQ \mW_c)\rangle \end{bmatrix} + \lambda \nabla^2 \regularizer(\vtheta) \vd$\\
      $t \gets \gamma / \vd^\top \mG \vd$ \label{line:t} \\
      $\vs \gets \vs + t \vd$ \\
      $\vr \gets \vr - t \mG \vd$ \\
      $\gamma_\text{new} = \vr^\top \vr$\\
      $\beta \gets \gamma_\text{new} / \gamma$ \\
      $\vd \gets \vr + \beta \vd$ \label{line:d} \\  
      $\gamma \gets \gamma_\text{new}$
    }
    \KwOut{$\vs$}
\end{algorithm}

\begin{algorithm}[t]
    \caption{An efficient implementation of Gauss-Newton method for solving \eqref{eq:prob}}
    \label{alg:GN-detailed}
  \DontPrintSemicolon
	\KwIn{$\sU=\{\dots, \vu_i, \dots\}, \sV=\{\dots, \vv_j, \dots\}, \obs, \tilde{\mP}, \tilde{\mQ}, \Wu, \Wv$.}
	\KwInit{\\
  \Indp
    Draw $\vtheta$ randomly \\
    $\fip{\tPc}{\tQc} \gets \fip{\tilde{\mP}^\top \Wu \tilde{\mP}}{\tilde{\mQ}^\top \Wv \tilde{\mQ}}$ \\
    $L(\vtheta), \mP, \mQ, \Pc, \Qc, \hPc, \hQc \gets \text{Alg. \ref{alg:funval-detailed}}(\vtheta, \sU, \sV, \obs, \tilde{\mP}, \tilde{\mQ}, \Wu, \Wv, \fip{\tPc}{\tQc})$
  }
  \Repeat{\text{stopping condition is satisfied}}{
    $\grad \gets \text{Alg. \ref{alg:grad-detailed}}(\vtheta, \obs, \tilde{\mP}, \tilde{\mQ}, \Wu, \Wv, \mP, \mQ, \Pc, \Qc, \hPc, \hQc)$ \\
    $\vs \gets \text{Alg. \ref{alg:cg-detailed}}(\vtheta, \obs, \grad, \Wu, \Wv, \mP, \mQ, \Pc, \Qc)$ \\
    $\delta \gets 1$ \\
    \While{line-search steps are within a limit}{
      $L(\vtheta + \delta \vs), \mP, \mQ, \Pc, \Qc, \hPc, \hQc \gets \text{Alg. \ref{alg:funval-detailed}}(\vtheta + \delta \vs, \sU, \sV, \obs, \tilde{\mP}, \tilde{\mQ}, \Wu, \Wv, \fip{\tPc}{\tQc})$ \\
      \lIf{
        $L(\vtheta + \delta \vs) \leq L(\vtheta) + \eta \delta \vs^\top \grad$
      }{\Break}
      $\delta \gets \delta/2$
    }
    $\vtheta \gets \vtheta + \delta \vs$
  }
\end{algorithm}

\subsection{Detailed Procedures of Gauss-Newton Method}
\label{sec:D_GN}
Details of the CG method involving a sequence of $\mG\vd$ operations are given in Algorithm \ref{alg:cg-detailed}.
The overall procedures of the Gauss-Newton method are summarized in Algorithm \ref{alg:GN-detailed}.

\end{document}